\documentclass[twoside,11pt]{article}

\usepackage[preprint]{jmlr2e}
\usepackage{xcolor}
\usepackage{multirow}
\usepackage{amsmath}
\newcommand{\Cov}{\mathrm{Cov}}
\newcommand{\Var}{\mathrm{Var}}
\newcommand{\tr}{\mathrm{tr}}
\usepackage{algorithmic}
\usepackage[ruled]{algorithm2e}
\usepackage{subfigure}
\newtheorem{assumption}[theorem]{Assumption}

\usepackage{lastpage}
\jmlrheading{25}{2024}{1-\pageref{LastPage}}{3/23; Revised
4/24}{5/24}{23-0379}{Yiling Xie and Xiaoming Huo}
\ShortHeadings{Adjusted Wasserstein Distributionally Robust Estimator}{Xie and Huo}
\firstpageno{1}

\begin{document}

\title{Adjusted Wasserstein Distributionally Robust Estimator in Statistical Learning}

\author{\name Yiling Xie \email yxie350@gatech.edu\\
\name Xiaoming Huo \email huo@gatech.edu \\
\addr School of Industrial and Systems Engineering\\
 Georgia Institute of Technology\\
Atlanta, Georgia, USA}
\editor{Po-Ling Loh}

\maketitle

\begin{abstract}%
We propose an {\it adjusted} Wasserstein distributionally robust estimator---based on a nonlinear transformation of the Wasserstein distributionally robust (WDRO) estimator in statistical learning.
The classic  WDRO estimator is asymptotically biased, while our adjusted WDRO estimator is asymptotically \textit{unbiased}, resulting in a smaller asymptotic mean squared error.
Further, under certain conditions, our proposed adjustment technique provides a general principle to de-bias asymptotically biased estimators. Specifically, we will investigate how the adjusted WDRO estimator is developed in the generalized linear model, including logistic regression, linear regression, and Poisson regression.
Numerical experiments demonstrate the favorable practical performance of the adjusted estimator over the classic one.
\end{abstract}


\section{Introduction}
 
  Wasserstein distributionally robust optimization (WDRO) has appeared as a promising tool to achieve ``robust" decision-making \citep{ mohajerin2018data,blanchet2019quantifying,gao2022distributionally}. 
WDRO has attracted intense research interest in the past few years. 
It is well-known that WDRO admits tractable reformulations \citep{mohajerin2018data} 
and has a powerful out-of-sample performance guarantee \citep{gao2022finite}.
People also have been actively exploring its applications in financial portfolio selection \citep{blanchet2022distributionally}, statistical learning \citep{chen2018robust,shafieezadeh2019regularization}, neural networks \citep{sinha2018certifying}, automatic control \citep{yang2020wasserstein}, transportation \citep{carlsson2018wasserstein}, and energy systems \citep{wang2018risk}, among many others.

WDRO can be applied in statistical learning \citep{chen2018robust,kuhn2019wasserstein,nguyen2022distributionally}.
In general, the statistical learning model can be written as the following optimization problem:
\[\min_{\beta\in B} \mathbb{E}_{P_{\ast}} \left[L(f(\mathbf{X},\beta),Y)\right],\]
where $\mathbf{X}\in\Omega\subset\mathbb{R}^d$ denotes the feature variable, $\Omega$ is a convex set, $Y$ denotes the response variable, $P_{\ast}$ is the true data-generating distribution of $(\mathbf{X},Y)$, $f(\cdot,\beta)$ is the hypothesis function parameterized by $\beta\in B\subset\mathbb{R}^d$, $B$ is a compact convex set, and $L$ is the loss function.
Considering the true data-generating distribution $P_{\ast}$ is usually unknown,  the empirical risk minimization can be applied to estimate the ground-truth hypothesis function $f(\cdot,\beta_\ast)$ parameterized by $\beta_{\ast}\not=\mathbf{0}$. 
However, the empirical risk minimization estimators are sensitive to perturbations and suffer from overfitting \citep{smith2006optimizer,shalev2014understanding}.
To obtain robust estimators with desirable generalization abilities, distributionally robust optimization is proposed, which minimizes the worst-case expected loss among an ambiguity set $\mathcal{U}$ of distributions. 
In this paper, we are interested in the Wasserstein ambiguity set, and then the resulting problem is the so-called  Wasserstein distributionally robust optimization.
 The Wasserstein ambiguity is defined as the ball centered at the empirical distribution $\mathbb{P}_n$ and contains all distributions close to $\mathbb{P}_n$ in the sense of the Wasserstein distance. 
 We denote the WDRO estimators---the solutions to the WDRO problem---by $\beta_n^{DRO}$.
 More details will be stated in Section \ref{adro}.

The asymptotic distribution of the WDRO estimator $\beta_n^{DRO}$ can be obtained under certain regularity conditions.
However, the associated convergence results imply that the WDRO estimator $\beta^{DRO}_n$ has an asymptotic bias.
From the perspective of parameter estimation, the asymptotic bias indicates an inaccurate estimation of the ground-truth parameter $\beta_{\ast}$.
Inspired by this phenomenon, we provide a general adjustment technique to de-bias the asymptotically biased estimators.
The asymptotic behavior of the asymptotically biased estimator under different transformations is also discussed.

We obtain the adjusted WDRO estimator, denoted by $\beta_n^{ADRO}$, by applying the proposed adjustment technique to the WDRO problem.
It will be shown that the adjusted WDRO estimator $\beta_n^{ADRO}$ could be computed exactly simply using the given samples and the value of the classic WDRO estimator $\beta_n^{DRO}$, making it convenient to apply the proposed technique.
Also, the existence and the asymptotic unbiasedness of the adjusted WDRO estimator $\beta_n^{ADRO}$ could be promised under mild conditions, enabling broad applications of the proposed technique.
In addition, since the proposed adjusted WDRO estimator $\beta^{ADRO}_n$ is transformed from the classic WDRO estimator $\beta^{DRO}_n$, the out-of-sample guarantee of the WDRO estimator $\beta^{DRO}_n$ could promise the generalization capacity of the proposed adjusted WDRO estimator $\beta^{ADRO}_n$.

Since the generalized linear model includes multiple widely-used regression models and is easy to interpret and implement, we will articulate how to apply the adjustment strategy in the setting of the generalized linear model, including linear regression, logistic regression, and Poisson regression.
Then, we carry out the numerical experiments in the generalized linear model.
Our numerical experiments illustrate that the proposed estimator $\beta_n^{ADRO}$ has a superior performance even if the sample size is not very large.


\subsection{Related Work}
We review the existing work related to the proposed adjusted WDRO estimator.
WDRO is broadly applied to solve parameter-estimation problems \citep{kuhn2019wasserstein,shafieezadeh2019regularization,aolaritei2022performance,nguyen2022distributionally}. 
Multiple algorithms have been developed \citep{li2019first,luo2019decomposition,blanchet2022optimal} and can be applied to compute the estimators in the WDRO framework.
While intense work focuses on adapting WDRO to different machine learning problems, deriving the tractable reformulations, and solving the WDRO problems efficiently, people have begun to investigate the statistical properties of WDRO estimators in recent few years, e.g., \cite{blanchet2021statistical,blanchet2022confidence,xie2024asymptotic}, evaluating the behavior of WDRO through the lens of statistics.
Notably, the asymptotic distribution of the WDRO estimator has been proven to be normal and has an asymptotic bias \citep{blanchet2022confidence}.
In this paper, we propose a nonlinear transformation to overcome this shortcoming.
It will be shown that the estimator obtained from the nonlinear transformation has an asymptotically smaller mean squared error, indicating the proposed estimator is more accurate in the asymptotic sense.
In the literature of WDRO, the generalization bounds, i.e., the upper confidence bounds on the out-of-sample loss, have been established to guarantee the out-of-sample performance of the WDRO estimator \citep{mohajerin2018data,shafieezadeh2019regularization,gao2022finite}.
Since the proposed adjusted WDRO estimator is transformed from the classic WDRO estimator, we can also develop the generalization bounds for the associated adjusted WDRO estimator. 
\subsection{Organization of this Paper}
The remainder of this paper is organized as follows.
In Section \ref{adjust}, we introduce the adjustment technique that could de-bias the general asymptotically biased estimators under certain conditions.
In Section \ref{basic}, we discuss the asymptotic behavior of the WDRO problem.
In Section \ref{adro}, we give the formulation of the adjusted WDRO estimator in statistical learning.
In Section \ref{glmdro}, we show how to develop the adjusted WDRO estimators in the generalized linear model.
Numerical experiments are conducted and analyzed in Section \ref{exp}.
The proofs are relegated to the appendix whenever possible.

\section{Adjustment Technique}\label{adjust}
In this section, we first discuss the properties of transformations on the asymptotically biased estimators, based on which we provide a general strategy to de-bias the asymptotically biased estimators under certain conditions. The proposed adjustment technique will be further illustrated in detail in the WDRO setting in Section \ref{adro}.

Suppose the estimator $\beta_n\in\mathbb{R}^d$ is obtained by the following parameter-estimation procedure: 
\begin{equation*}\label{procedure}\beta_n\in\arg\min_{\beta} l(\mathbb{P}_n,\beta),\end{equation*}
where $l$ is the loss and depends on the empirical distribution $\mathbb{P}_n$ and parameter $\beta$.
Also, suppose that the estimator $\beta_n$  has the following convergence in distribution:
\begin{equation}\label{general}\sqrt{n}(\beta_n-\beta_\ast)\Rightarrow\mathcal{N}(f(\beta_\ast),D),\end{equation}
where $\Rightarrow$ means ``converge in distribution'',  $D\in\mathbb{R}^{d\times d}$ is the asymptotic covariance matrix, $f:\mathbb{R}^d\mapsto\mathbb{R}^d$ and $\beta_\ast\in\mathbb{R}^d$ is the ground-truth parameter. We focus on the scenario when $f\not=0$.

For the estimator $\beta_n$ with the limiting distribution in \eqref{general}, our goal is to look for some (deterministic) transformation $\phi_n$ to obtain a more accurate estimation of $\beta_\ast$ in the asymptotic sense. 
The following proposition states that the ``best'' transformations have a unique formulation.
\begin{proposition}\label{uniquetransformation}
Suppose  $\beta_n$ is an estimator of ground-truth parameter $\beta_\ast$ and has the following convergence in distribution:
\[\sqrt{n}(\beta_n-\beta_\ast)\Rightarrow\mathcal{N}(f(\beta_\ast),D),\]
 where $f$  is differentiable at some neighborhood $\mathcal{B}(\beta_\ast)$ of $\beta_\ast$.
 Assume the transformation $\phi_n$ is differentiable at $\mathcal{B}(\beta_\ast)$ and  satisfies $\phi_n(\beta)\to \phi(\beta)$ and $\phi_n^{\prime}(\beta)\to \phi^{\prime}(\beta)$ for every $\beta$ in $\mathcal{B}(\beta_\ast)$, where $\phi$ is differentiable, and $\phi_n^{\prime}$ and $\phi^\prime$ are the gradients of $\phi_n$ and $\phi$.
Under this assumption, the least asymptotic mean squared error of $\phi_n(\beta_n)$ is $\tr(D)$, which can be achieved if and only if the transformation $\phi_n$ has the following formulation
\begin{equation}\label{best}
\phi_n(\beta)=\beta-\frac{1}{\sqrt{n}}g(\beta)+o\left(\frac{1}{\sqrt{n}}\right),
\end{equation}
where $g$ is some differentiable function at $\mathcal{B}(\beta_\ast)$ satisfying $g(\beta_\ast)=f(\beta_\ast)$,
resulting in the following convergence in distribution: 
\[\sqrt{n}(\phi_n(\beta_n)-\beta_\ast)\Rightarrow\mathcal{N}(0,D).\]
\end{proposition}

Proposition \ref{uniquetransformation} demonstrates that for the asymptotically biased estimator $\beta_n$,  the transformation $\phi_n$ should take the formulation \eqref{best} to achieve the least asymptotic mean squared error $\tr(D)$. 
Meanwhile, the resulting estimator $\phi_n(\beta_n)$ is asymptotically unbiased. 

The transformation $\phi_n$ in the formulation \eqref{best} is desirable, and one can simply let $g=f$ to define the transformation $\phi_n$ in \eqref{best}.
However,  the function $f$  is usually unknown. 
For example, in the limiting distribution of the WDRO estimator, $f$ depends on the unknown ground-truth data-generating distribution. 
In this regard, the function $f$ should be approximated accordingly.

Suppose we have a sequence of (stochastic) functions $f_n$ to approximate the function $f$. Our adjustment transformation is defined in terms of $f_n$ and based on the formulation of $\phi_n$ shown in \eqref{best}.
Certain conditions should be imposed to $f_n$ to promise that the estimator obtained by our adjustment transformation is asymptotically unbiased and could have the asymptotic mean squared error $\tr(D)$.
More details are described in Assumption \ref{assumefn} and Theorem \ref{generaltheorem}.

Before introducing Theorem \ref{generaltheorem}, we state our assumptions of functions $f_n$.
\begin{assumption}\label{assumefn}
Given function $f$, $f_n$ and $\beta_\ast$, we assume that 
    \begin{itemize}
    \item The function $f_n$ is differentiable at some neighborhood $\mathcal{B}(\beta_\ast)$ of $\beta_\ast$.
    \item The sequence $\sup_{\beta\in \mathcal{B}(\beta_\ast)}\Vert f^\prime_n(\beta)\Vert$ is bounded in probability.
    \item $f_n(\beta_\ast)\to_p f(\beta_\ast)$, where $\to_p$ means ``converge in probability''.
\end{itemize}
\end{assumption}
Equipped with Assumption \ref{assumefn}, we give our main result in the following theorem.
\begin{theorem}[Adjustement Technique]\label{generaltheorem}
Suppose  $\beta_n$ is an estimator of ground-truth parameter $\beta_\ast$ and has the following convergence in distribution:
\[\sqrt{n}(\beta_n-\beta_\ast)\Rightarrow\mathcal{N}(f(\beta_\ast),D),\]
 where $f$  is differentiable at some neighborhood $\mathcal{B}(\beta_\ast)$ of $\beta_\ast$.
If we have the function $f_n$ satisfying 
Assumption \ref{assumefn} and the transformation $\mathcal{A}_n$ defined by
\[\mathcal{A}_n(\beta_n)=\beta_n-\frac{1}{\sqrt{n}}f_n(\beta_n),\]
then we have that \begin{equation}\label{generaladjusted}\sqrt{n}\left(\mathcal{A}_n(\beta_n)-\beta_\ast\right)\Rightarrow\mathcal{N}(0,D).\end{equation}
\end{theorem}

The convergence \eqref{generaladjusted} in Theorem \ref{generaltheorem} demonstrates that the proposed adjusted estimator $\mathcal{A}_n(\beta_n)$ is asymptotically unbiased and the asymptotic covariance matrix remains unchanged, resulting in a smaller asymptotic means square error $\tr(D)$, which is the least asymptotic mean squared error stated in Proposition \ref{uniquetransformation}.
In this regard, to de-bias the asymptotically biased estimators, one only needs to have a sequence of functions $f_n$ satisfying Assumption \ref{assumefn}.
\subsection{Sequential Delta Method}
Notice that the transformations $\phi_n$ discussed in Proposition \ref{uniquetransformation} depend on $n$. In this way, when we discuss the asymptotic distribution of $\phi_n(\beta_n)$, the classic delta method is not applicable. To resolve this issue, we have developed a sequential delta method based on the extended continuous mapping theorem, seeing Theorem 1.11.1 in \cite{van1996weak}. The sequential delta method may have an independent research interest, so we state it in the following theorem.
\begin{theorem}[Sequential Delta Method]\label{extendeddeltamethod}
    Let $\phi_n$ and $\phi:\mathbb{D}\subset\mathbb{R}^d\mapsto \mathbb{R}^d$ be functions defined on a subset of $\mathbb{R}^d$. Suppose $\phi_n$ and $\phi$ are differentiable at the neighborhood $\mathcal{B}(\vartheta)\subset\mathbb{D}$ of $\vartheta\in \mathbb{D}$, and $\phi_{n}(\theta)\to \phi(\theta)$ and $\phi^\prime_{n}(\theta)\to \phi^\prime(\theta)$
     hold for every $\theta\in\mathcal{B}(\vartheta)$, where $\phi^\prime$ and $\phi^\prime_n$ are gradients of the functions $\phi$ and $\phi_n$.
    Let $T_n$ be random vectors taking their values in $\mathbb{D}$. 
    If $r_n(T_n-\vartheta)\Rightarrow \mathcal{N}(\mu,\Sigma)$ for numbers $r_n\to\infty$, then we have that \[r_n(\phi_n(T_n)-\phi_n(\vartheta))\Rightarrow \mathcal{N}(\phi^\prime(\vartheta) \mu ,\phi^\prime(\vartheta)\Sigma{\phi^\prime(\vartheta)}^\top).\]
\end{theorem}
\section{WDRO Problem}\label{basic}
This section discusses the problem formulation of WDRO and gives the asymptotic distribution of the WDRO estimator.
\subsection{Problem Formulation}
The WDRO problem can be written as
\begin{equation}\label{formal}\beta^{DRO}_n\in\arg\min_{\beta\in B}\sup_{P\in \mathcal{U}_{\rho_n}(\mathbb{P}_n)} \mathbb{E}_{P} \left[L(f(\mathbf{X},\beta),Y)\right],\end{equation}
 where the feature variable $\mathbf{X}$ belongs to the convex set $\Omega\subset \mathbb{R}^d$, the response variable $Y$ can be continuous or discrete, $f$ is the hypothesis function parametrized by $\beta\in B\subset\mathbb{R}^d$, $B$ is a compact convex set, $\mathcal{U}_{\rho_n}(\mathbb{P}_n)$ is the Wasserstein uncertainty set, and $L$ is the loss function. The Wasserstein uncertainty set is defined by
 \begin{equation}\label{uncertainty}\mathcal{U}_{\rho_n}(\mathbb{P}_n)=\{P: W_p(P,\mathbb{P}_n)\leq \rho_n\},\end{equation}
where $\mathbb{P}_n$ is the empirical distribution of the samples $\{(\mathbf{X}_1,Y_1), (\mathbf{X}_2,Y_2),..., (\mathbf{X}_n,Y_n)\}$ generated by true data-generating distribution $P_{\ast}$,
\[W_p(P,\mathbb{P}_n)=\left(\inf_{\gamma\in\Gamma(P,\mathbb{P}_n)}\left\{\int_{Z^2} d^p(z,z^{\prime}) d\gamma(z,z^{\prime})\right\}\right)^{1/p},\]
$\Gamma(P,\mathbb{P}_n)$ is the set of  distributions with marginals $P$  and  $\mathbb{P}_n$,
$d$ is some metric in space $Z=\mathbf{X}\times Y$,
and $W_p(P_1,P_2)$ is the so-called $p$-Wasserstein distance.

\subsection{Asymptotic Distribution of the WDRO Estimator}\label{section2.1}
In this subsection, we study the asymptotic distribution of the WDRO estimator in the supervised statistical learning.

\cite{blanchet2022confidence} have derived the asymptotic distribution of the WDRO estimator in the unsupervised learning setting. 
In our study, we first let the cost function be infinite if the response variables are different and then adapt the asymptotic distribution of the WDRO estimator to the supervised statistical learning setting.

To adapt the results, we should specify the hyperparameters of the Wasserstein uncertainty set and clarify some regularity conditions, which should be satisfied for the loss function $L$ and the underlying data-generating distribution $P_{\ast}$ of $(\mathbf{X},Y)$.
\begin{assumption}\label{assume1}
The hyperparameters of the Wasserstein uncertainty set $\mathcal{U}_{\rho_n}(\mathbb{P}_n)$ in \eqref{uncertainty} are prescribed as follows,
\begin{itemize}
\item $\rho_n=\tau/\sqrt{n}$, $\tau>0$,
\item $p=2$,
\item $d\left((\mathbf{x}_1,y_1),(\mathbf{x}_2,y_2)\right)=
\begin{cases}
\Vert\mathbf{x}_1-\mathbf{x}_2\Vert_2&y_1=y_2\\
\infty&y_1\not=y_2
\end{cases}$.
\end{itemize}
\end{assumption}  
\begin{remark}
We justify the choices of hyperparameter in Assumption \ref{assume1} as follows,
\begin{itemize}
    \item 
We choose the radius to be of the square-root order $\mathcal{O}(1/\sqrt{n})$ because the powerful out-of-sample performance guarantee can be proved \citep{gao2022distributionally}, and the confidence region can be constructed \citep{blanchet2022confidence} with the square-root order.
\item
We choose the 2-Wasserstein distance since the 2-Wasserstein distance applies to the quadratic loss, and the associated WDRO problem could be solved by iterative algorithms \citep{blanchet2022optimal}.
\item
The distance function $d$ is infinite when $y_1\not=y_2$, admitting distributional ambiguities only with respect to the feature variable $\mathbf{X}$.
 In the classification problem, the distance function $d$ can be applied to tasks where the samples are correctly labeled \citep{gao2017distributional}.
 In the regression problem, the distance function $d$ can help recover several popular regularized estimators, including square-root LASSO estimator \citep{blanchet2019robust,shafieezadeh2019regularization}.
 \end{itemize}
\end{remark}
\begin{assumption}\label{assumeloss}
The loss function $L(f(\mathbf{x},\beta),y)$ satisfies:
\begin{itemize}
    \item[a.]  The loss function $L(f(\mathbf{x},\beta),y)$  is twice continuously differentiable w.r.t. $\mathbf{x}$ and $\beta$. 
    \item [b.] For each variable $\mathbf{x}\in\Omega$ and $y$, the loss function $L(f(\mathbf{x},\beta),y)$ is convex w.r.t. $\beta$. 
    \item[c.] For each parameter $\beta\in B$ and variable y, the function $\left\Vert\frac{\partial^2 L(f(\mathbf{x},\beta),y)}{\partial \mathbf{x}^2}\right\Vert_2$ is uniformly continuous w.r.t. $\mathbf{x}$ and  uniformly bounded by a continuous function $M(\beta)$.
\end{itemize}
\end{assumption}
\begin{assumption}\label{assumedistr}
The underlying data-generating distribution $P_{\ast}$ of $(\mathbf{X},Y)$ satisfies:
\begin{itemize}
    \item[a.] There exists $\beta_{\ast}\in B^{\circ}$, where $B^{\circ}$ means the interior of $B$, satisfying
    \[\mathbb{E}_{P_{\ast}}\left[\frac{\partial L(f(\mathbf{X},\beta),Y)}{\partial\beta}\right]\Bigg\vert_{\beta=\beta_{\ast}}=\mathbf{0},\]
    and the inequalities
    \begin{equation}\label{matrixD}C(\beta_\ast):=\mathbb{E}_{P_{\ast}}\left[\frac{\partial^2 L(f(\mathbf{X},\beta),Y)}{\partial \beta^2}\right]\Bigg\vert_{\beta=\beta_{\ast}}\succ 0,\end{equation}
    \[\mathbb{E}_{P_{\ast}} \left[\left\Vert \frac{\partial L(f(\mathbf{X},\beta),Y)}{\partial \beta}\right\Vert_2^2\right]\Bigg\vert_{\beta=\beta_{\ast}}<\infty\] 
    hold,  where $C(\beta_\ast)\succ0$ means the matrix $C(\beta_\ast)$ is a positive definite matrix.
    \item[b.] $P_{\ast}$ is non-degenerate in the sense that  
    \[P_{\ast}\left(\frac{\partial L(f(\mathbf{X},\beta),Y)}{\partial \mathbf{X}}\not =\mathbf{0}\right)\bigg\vert_{\beta=\beta_{\ast}}>0,\]
    \[\mathbb{E}_{P_{\ast}}\left[\frac{\partial ^2L(f(\mathbf{X},\beta),Y)}{\partial \mathbf{X}\partial \beta}\left(\frac{\partial ^2L(f(\mathbf{X},\beta),Y)}{\partial \mathbf{X}\partial \beta}\right)^{\top}\right]\Bigg\vert_{\beta=\beta_{\ast}}\succ 0,\]
    where $\frac{\partial ^2 L}{\partial \mathbf{x}\partial \beta}$ means taking the gradient first w.r.t. $\beta$ and then w.r.t. $\mathbf{x}$.
\end{itemize}
\end{assumption}

\begin{figure}
    \centering
    \begin{minipage}{0.5\textwidth}
        \centering
        \includegraphics[width=\textwidth]{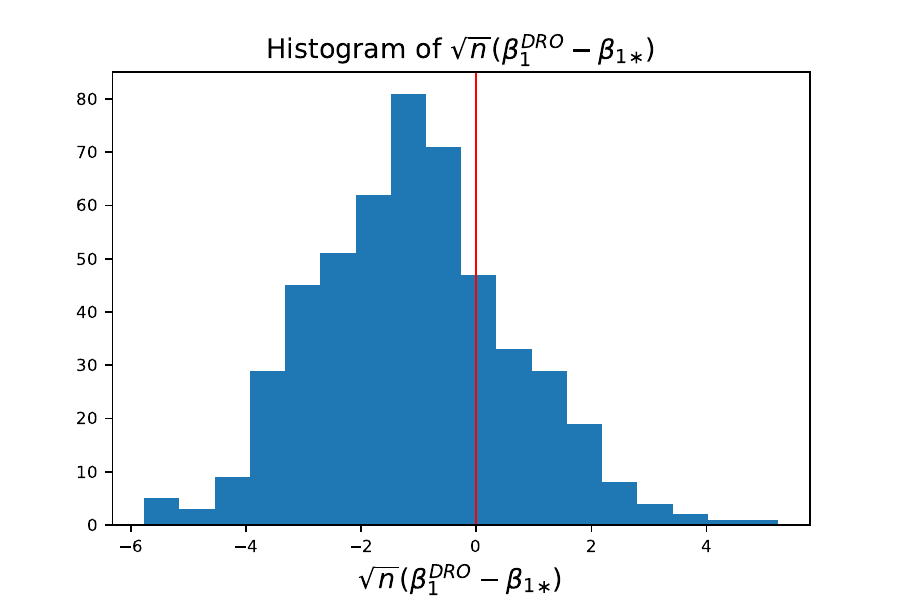} 
        
    \end{minipage}\hfill
    \begin{minipage}{0.5\textwidth}
        \centering
        \includegraphics[width=\textwidth]{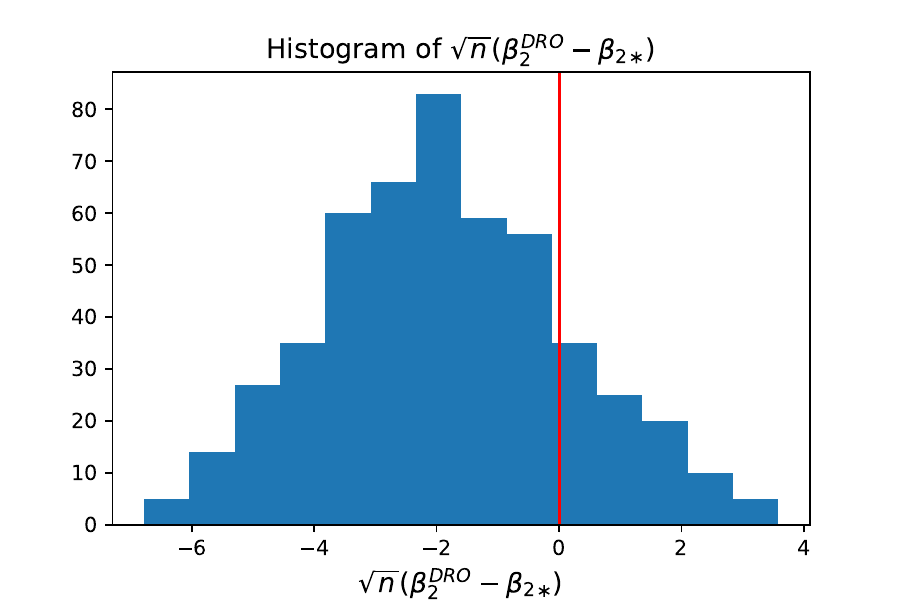} 
    \end{minipage}
    \caption{Histogram of $\beta_n^{DRO}$}
    \label{hist}
\end{figure}

Next, we obtain the associated convergence of the WDRO estimator $\beta_n^{DRO}$ in problem \eqref{formal} under Assumption \ref{assume1}, \ref{assumeloss}, and \ref{assumedistr}, which is shown in the following theorem.
\begin{theorem}[Extension of Theorem 1 in \cite{blanchet2022confidence}]
\label{theorem1}
Suppose that Assumption \ref{assume1}, \ref{assumeloss} and \ref{assumedistr} are satisfied, $\Omega=\mathbb{R}^d$ and $\mathbb{E}_{P_\ast}\left[\Vert\mathbf{X}\Vert_2^2\right]<\infty$, the WDRO estimator $\beta_n^{DRO}$ in problem \eqref{formal} has the following convergence in distribution:
\begin{equation}\label{con}\sqrt{n}(\beta_{n}^{\text{DRO}}-\beta_{\ast})\Rightarrow\mathcal{N}\left(-C(\beta_\ast)^{-1}H\left(\beta_{\ast}\right),D(\beta_\ast)\right),\end{equation}
where 
\begin{equation}\label{orginH}H(\beta_{\ast})=\tau\frac{\partial \sqrt{\mathbb{E}_{P_\ast}\left[\left\Vert \frac{\partial L(f(\mathbf{X},\beta),Y)}{\partial \mathbf{X}}\right\Vert_2^2\right] }}{\partial \beta}\Bigg\vert_{\beta=\beta_{\ast}},\end{equation}
 $\tau$ is the coefficient in the Wasserstein radius $\rho_n=\tau/\sqrt{n}$,
\begin{equation}\label{covariancematrix}D(\beta_\ast)=C(\beta_\ast)^{-1}\Cov\left(\frac{\partial L(f(\mathbf{X},\beta),Y)}{\partial \beta}\right)\bigg\vert_{\beta=\beta_{\ast}}C(\beta_\ast)^{-1},\end{equation}
and $C(\beta_\ast)$ is defined in \eqref{matrixD}.
\end{theorem}
\begin{remark}
The assumption $\Omega=\mathbb{R}^d$ could be relaxed. If $\Omega$ is compact and could be expressed as $\Omega=\{\mathbf{x}\in\mathbb{R}^d: A\mathbf{x}\leq b\}$,  where $A$ is an $l\times d$ matrix with linearly independent rows and $b\in\mathbb{R}^l$, and $\mathbf{X}$ has a probability density which is absolutely continuous w.r.t. Lebesgue measure, then the convergence \eqref{con} still holds. This claim can be seen in Section 6 in \cite{blanchet2022confidence}.
\end{remark}
\begin{remark}[Finite Sample Size]
We investigate the empirical distribution of $\beta_n^{DRO}$ when $n$ is not very large. The WDRO esitmator $\beta_n^{DRO}$ is computed in the logistic regression model when $n=200$, and we plot the histograms of $\sqrt{n}(\beta_n^{DRO}-\beta_\ast)$ in Figure \ref{hist}. Two dimensions of  $\beta_n^{DRO}$ are plotted separately.  We conclude from Figure \ref{hist} that $\beta_n^{DRO}$ is approximately normally distributed with a nonzero mean, as asymptotic convergence \eqref{con} suggested. 
We further apply the Shapiro–Wilk test and the test result supports our claim that $\beta_n^{DRO}$ is approximately normally distributed even though the sample size is not very large, indicating that the asymptotic behavior of $\beta_n^{DRO}$ ``comes early''.
Therefore, making the bias in asymptotic convergence \eqref{con} disappear is meaningful in the sense of both asymptotic and finite sample size.
\end{remark}

Theorem \ref{theorem1} indicates that the term $\sqrt{n}(\beta_{n}^{DRO}-\beta_\ast)$ converges in distribution to a normal distribution with nonzero mean $-C(\beta_\ast)^{-1}H(\beta_{\ast})$. 
Recall that we perturb the samples to achieve robustification.
As explained in \cite{blanchet2021statistical}, the bias term $-C(\beta_\ast)^{-1}H(\beta_{\ast})$ could be understood as pushing towards solutions with less variation resulting from data perturbation.
However, this nonzero bias term may imply that the WDRO estimator is not an accurate estimator for the ground-truth parameter $\beta_{\ast}$.
We may consider transforming the WDRO estimator $\beta_n^{DRO}$ to remove the bias term using the adjustment technique mentioned in Section \ref{adjust}.

\section{Proposed Adjusted WDRO Estimator}\label{adro}
This section introduces the formal formulation of our adjusted WDRO estimator and investigates the relevant properties, including unbiasedness, possible simplification, and the out-of-sample guarantee.

\subsection{Definition and Existence}
The adjusted WDRO estimator is based on the asymptotic distribution obtained in Section \ref{section2.1} and the adjustment technique introduced in Section \ref{adjust}.
Recall the WDRO estimator has the following convergence:
\[\sqrt{n}(\beta_{n}^{\text{DRO}}-\beta_{\ast})\Rightarrow\mathcal{N}\left(-C(\beta_\ast)^{-1}H\left(\beta_{\ast}\right),D(\beta_\ast)\right),\]
where 
\[C(\beta_\ast)=\mathbb{E}_{P_{\ast}}\left[\frac{\partial^2 L(f(\mathbf{X},\beta),Y)}{\partial \beta^2}\right]\Bigg\vert_{\beta=\beta_{\ast}}, \quad H(\beta_{\ast})=\tau\frac{\partial \sqrt{\mathbb{E}_{P_\ast}\left[\left\Vert \frac{\partial L(f(\mathbf{X},\beta),Y)}{\partial \mathbf{X}}\right\Vert_2^2\right] }}{\partial \beta}\Bigg\vert_{\beta=\beta_{\ast}}.\]

Notice that the asymptotic bias $f(\beta_\ast)=-C(\beta_\ast)^{-1}H\left(\beta_{\ast}\right)$ depends on the unknown underlying data-generating distribution $P_\ast$, but we can use the associated empirical distribution to approximate $f$. Applying the adjusted technique proposed in Theorem \ref{generaltheorem}, we define the adjusted WDRO estimator in the following.
\begin{definition}[Adjusted WDRO Estimator]\label{def}
In the WDRO problem \eqref{formal}, under Assumption \ref{assume1}, \ref{assumeloss}, and \ref{assumedistr}, the adjusted WDRO estimator is defined by
\begin{equation}\label{adrodefinition}\beta_n^{ADRO}=\mathcal{A}_n(\beta_n^{DRO}),\end{equation}
where
\[\mathcal{A}_n(\mathbf{z})=\mathbf{z}+\frac{C_n(\mathbf{z})^{-1}H_n(\mathbf{z})}{\sqrt{n}},\]
\begin{equation}\label{functionH}H_n(\mathbf{z})=\tau\frac{\partial \sqrt{\mathbb{E}_{\mathbb{P}_n}\left[\left\Vert \frac{\partial L(f(\mathbf{X},\beta),Y)}{\partial \mathbf{X}}\right\Vert_2^2 \right]}}{\partial \beta}\Bigg\vert_{\beta=\mathbf{z}},\end{equation}
\begin{equation}\label{functionC}C_n(\mathbf{z})=\mathbb{E}_{\mathbb{P}_n}\left[\frac{\partial^2 L(f(\mathbf{X},\beta),Y)}{\partial \beta^2}\right]\Bigg\vert_{\beta=\mathbf{z}}.\end{equation}
\end{definition}

To promise the existence of the adjusted WDRO estimator, we need additional conditions to let the matrix $C_n(\beta_n^{DRO})$ be invertible and the vector $H_n(\beta_n^{DRO})$ be well-defined. The conditions are shown in the following proposition.
\begin{proposition}[Existence of Adjusted WDRO Estimator I]\label{conditions}
    In the WDRO problem \eqref{formal}, under Assumption \ref{assume1}, \ref{assumeloss}, and \ref{assumedistr}, for the empirical distribution $\mathbb{P}_n$, the loss function $L(f(\mathbf{x},\beta),y)$ and the WDRO estimator $\beta_n^{DRO}$, if
    \[\mathbb{P}_n\left( \left\Vert \frac{\partial L(f(\mathbf{X},\beta),Y)}{\partial \mathbf{X}}\right\Vert_2^2\not=0\right)\Bigg\vert_{ \beta=\beta_n^{DRO}}>0,  \quad \mathbb{E}_{\mathbb{P}_n}\left[\frac{\partial^2 L(f(\mathbf{X},\beta),Y)}{\partial \beta^2}\right]\Bigg\vert_{\beta=\beta_n^{DRO}}\succ 0 \]
    hold, then the adjusted WDRO estimator $\beta_n^{ADRO}$ defined in \eqref{adrodefinition} exists. 
\end{proposition}

If the hypothesis function is linear, i.e., $f(\mathbf{x},\beta)=\langle \mathbf{x},\beta\rangle$, the existence conditions demonstrated in Proposition \ref{conditions} could be further simplified as shown in the following proposition.
\begin{proposition}[Existence of Adjusted WDRO Estimator II]\label{conditions2}
    In the WDRO problem \eqref{formal}, under Assumption \ref{assume1}, \ref{assumeloss}, and \ref{assumedistr}, for the empirical distribution $\mathbb{P}_n$, the loss function $L(\langle\mathbf{x},\beta\rangle,y)$ and the WDRO estimator $\beta_n^{DRO}$, if \[\beta_n^{DRO}\not=\mathbf{0},\quad \frac{\partial^2 L(f,y)}{\partial f^2}>0,\quad \mathbb{P}_n\left( \frac{\partial L(\langle\mathbf{X},\beta_n^{DRO}\rangle,Y)}{\partial f}\not=0\right)>0,\] hold,   where $\frac{\partial L}{\partial f}$ means taking the gradient of $L$ w.r.t. the first argument, and there does not exist nonzero vector $\alpha$ such that $\mathbb{P}_n(\alpha^{\top}\mathbf{X}=0)=1$, then the adjusted WDRO estimator $\beta_n^{ADRO}$ defined in \eqref{adrodefinition} exists.
\end{proposition}

The conditions in Proposition \ref{conditions} and \ref{conditions2} are mild. For example, for the nonzero WDRO estimator $\beta_n^{DRO}$ and non-degenerate loss $L$ with positive second-order derivative, if the feature variable $\mathbf{X}$ does not lie in any linear subspace of $\mathbb{R}^d$, the conditions in Proposition \ref{conditions2} can hold.
One may check that the existence conditions could be satisfied by multiple statistical models, including linear regression and logistic regression, among many others.
\subsection{Simplification of the Adjusted WDRO Estimator}
In this subsection, we discuss under which conditions the expression of the adjusted WDRO estimator $\beta_n^{ADRO}$ could be further simplified.

Recall that, in the definition of the adjusted WDRO estimator, seeing Definition \ref{def}, the term $H_n(\mathbf{z})$ appears complicated at first glance. The following proposition shows that the function $H_n(\mathbf{z})$ can be simplified under certain conditions.
\begin{proposition}[Simplification]\label{sufficient2}
    If the hypothesis function in problem \eqref{formal} is a linear function, i.e., $f(\mathbf{x},\beta)=\langle \mathbf{x},\beta\rangle$, and the equation 
    \begin{equation}\label{sufficient2eq}\mathbb{E}_{P_\ast}\left[ \frac{\partial L(\langle\mathbf{X},\beta\rangle,Y)}{\partial f}\frac{\partial L^2(\langle\mathbf{X},\beta\rangle,Y)}{\partial f^2}\mathbf{X}\right]\Bigg\vert_{\beta=\beta_\ast}=0\end{equation}
    holds, then the function $H(\beta_\ast)$ defined in \eqref{orginH} can be rewritten as  
\[H(\beta_\ast)=\tau\sqrt{\mathbb{E}_{P_\ast}\left[\left( \frac{\partial L(\langle\mathbf{X},\beta\rangle,Y)}{\partial f}\right)^2\right]}\Bigg\vert_{\beta=\beta_\ast}\frac{\beta_\ast}{{\Vert\beta_\ast\Vert_2}}.\] 
\end{proposition}

Proposition \ref{sufficient2} implies that the linearity of the hypothesis function and the equation \eqref{sufficient2eq} can promise that $H(\mathbf{z})$ is a rescaling of $\mathbf{z}$. The associated function $H_n(\mathbf{z})$ is defined by
\[H_n(\mathbf{z})=\tau\sqrt{\mathbb{E}_{\mathbb{P}_n}\left[\left( \frac{\partial L(\langle\mathbf{X},\beta\rangle,Y)}{\partial f}\right)^2\right]}\Bigg\vert_{\beta=\mathbf{z}}\frac{\mathbf{z}}{{\Vert\mathbf{z}\Vert_2}}.\] 
In this way, the expression of the adjusted WDRO estimator could be simplified. 
In particular, the conditions in Proposition \ref{sufficient2} can be satisfied by multiple statistical models, e.g., linear regression, logistic regression, and Poisson regression. The details can be found in Section \ref{glmdro}.
\subsection{Asymptotically Unbiased}
We establish the asymptotic distribution of the adjusted WDRO estimator $\beta_n^{ADRO}$.

\begin{theorem}[Unbiasedness]\label{theorem2}
Under Assumption \ref{assume1}, \ref{assumeloss}, and \ref{assumedistr}, if the adjusted WDRO estimator $\beta_n^{ADRO}$ defined in \eqref{adrodefinition} exists,
and $\frac{\partial L(f(\mathbf{x},\beta),y)}{\partial \mathbf{x}\partial \beta}$, $\frac{\partial^2 L(f(\mathbf{x},\beta),y)}{\partial \beta^2}$ are continuously differentiable w.r.t. $\beta$, then the adjusted WDRO estimator $\beta_n^{ADRO}$ converges in distribution:
\begin{equation*}\sqrt{n}(\beta_{n}^{ADRO}-\beta_{\ast})\Rightarrow\mathcal{N}(0,D(\beta_\ast)),\end{equation*} 
where $D(\beta_\ast)$ is defined in \eqref{covariancematrix}.
\end{theorem}

Theorem \ref{theorem2} indicates that our proposed estimator $\beta^{ADRO}_n$ is asymptotically unbiased
and the asymptotic mean squared error is $\tr(D(\beta_\ast))$.
Recall the asymptotic distribution of the classic WDRO estimator $\beta_n^{DRO}$ is 
\[\sqrt{n}(\beta_{n}^{\text{DRO}}-\beta_{\ast})\Rightarrow\mathcal{N}\left(-C(\beta_\ast)^{-1}H\left(\beta_{\ast}\right),D(\beta_\ast)\right),\]
indicating that the asymptotic mean squared error of the classic WDRO estimator $\beta_n^{DRO}$ is $\tr(D(\beta_\ast))+f(\beta_\ast)^\top f(\beta_\ast)$, where $f(\beta_\ast) = -C(\beta_\ast)^{-1}H(\beta_\ast)$ might not be zero. In this way, our proposed estimator has a smaller asymptotic mean squared error.
\subsection{Out-of-sample Performance Guarantee}
This subsection discusses the out-of-sample performance guarantee for the adjusted WDRO estimator $\beta_n^{ADRO}$.

Informally, the out-of-sample performance guarantee for the WDRO estimator $\beta_n^{DRO}$ reads that, with a high probability, the following inequality holds:
\begin{equation}\label{outofsample}
\mathbb{E}_{P_{\ast}}\left[ L(f(\mathbf{X},\beta_n^{DRO}),Y)\right]\leq\sup_{P\in \mathcal{U}_{\rho_n}(\mathbb{P}_n)} \mathbb{E}_{P} \left[L(f(\mathbf{X},\beta_n^{DRO}),Y)\right]+\epsilon_n, \end{equation}
where the left-hand side is the generalization error of $\beta_n^{DRO}$, and the first term on the right-hand side is called Wasserstein robust loss of $\beta_n^{DRO}$. Inequality \eqref{outofsample} implies that the ground-truth loss of $\beta_n^{DRO}$ is upper bounded by the Wasserstein robust loss up to a higher order residual $\epsilon_n$.

Recall that our proposed adjusted estimator $\beta_n^{ADRO}$ is transformed from the WDRO estimator $\beta_n^{DRO}$. As the  WDRO estimator $\beta_n^{DRO}$ enjoys the out-of-sample performance guarantee \eqref{outofsample}, similar arguments can be established towards the adjusted WDRO estimator $\beta_n^{ADRO}$. 
\begin{corollary}[Performance Guarantee]\label{coro1}

Suppose the generalization bound \eqref{outofsample} holds for the WDRO estimator $\beta_n^{DRO}$ for some residual term $\epsilon_n$ with probability $1-\alpha$.
If the loss function $L(f(\mathbf{x},\beta),y)$ is $h$-Lipschitz continuous w.r.t. $\beta$, and the adjusted WDRO estimator $\beta_n^{ADRO}$ exists, then the following inequality:
\begin{equation*}
\begin{aligned}
\mathbb{E}_{P_{\ast}}\left[ L(f(\mathbf{X},\beta_n^{ADRO}),Y)\right]\leq\sup_{P\in \mathcal{U}_{\rho_n}(\mathbb{P}_n)} \mathbb{E}_{P} \left[L(f(\mathbf{X},\beta_n^{ADRO}),Y)\right]
+ \frac{h}{\sqrt{n}}\mathcal{R}_n+\epsilon_n, 
\end{aligned}
\end{equation*}
where $ \mathcal{R}_n=\mathbb{E}_{P_\ast}\left[\Vert C_n(\beta_n^{DRO})^{-1}H_n(\beta_n^{DRO})\Vert_2\right]+ \sup_{P\in\mathcal{U}_{\rho_n}(\mathbb{P}_n)}\mathbb{E}_P\left[\Vert C_n(\beta_n^{DRO})^{-1}H_n(\beta_n^{DRO})\Vert_2\right]$, holds with probability $1-\alpha$.
\end{corollary}

Notably, \cite{gao2022finite} derives the generalization bound based on a novel variance-based concentration inequality for the empirical loss for the radius of the order $\mathcal{O}(1/\sqrt{n})$, where $\epsilon_n=\widetilde{\mathcal{O}}(1/n)$  ($\widetilde{\mathcal{O}}$ is used to suppress the logarithmic dependence). In this sense, Corollary \ref{coro1} indicates that the generalization error of the adjusted WDRO estimator $\beta_n^{ADRO}$ can be upper bounded by the Wasserstein robust loss of the adjusted WDRO estimator $\beta_n^{ADRO}$ up to a new residual term, $h\mathcal{R}_n/\sqrt{n} +\epsilon_n$, which is of order $\mathcal{O}(1/\sqrt{n})$.
The new residual order of the out-of-sample guarantee for the adjusted WDRO estimator may have a lower order than that of the classic WDRO estimator shown in \cite{gao2022finite}. To further improve the residual order for the adjusted WDRO estimator could be considered as our future work.
\section{Adjusted WDRO in the Generalized Linear Model}\label{glmdro}
In this section, the generalized linear model is considered since several well-known regression models can be covered, including logistic regression, Poisson regression, and linear regression.
We introduce how to develop the associated adjusted WDRO estimators in the generalized linear model.
\subsection{Formulation of the Generalized Linear Model}
In the generalized linear model, the response variable $Y$ is generated from a particular distribution from the exponential family, including the Bernoulli distribution on $Y\in\{-1,1\}$ in the logistic regression, the Poisson distribution on $Y\in\{0,1,2,...\}$ in the Poisson regression, the normal distribution on $Y\in\mathbb{R}$ in the linear regression, etc.
The expectation of the response variable $Y$ conditional on the feature variable $\mathbf{X}$ is determined by the link function.
With a little abuse of notation, if we denote the nonzero ground-truth parameter by $\beta_\ast$ and the link function by $G$, we have $G(\mathbb{E}[Y|\mathbf{X}=\mathbf{x}])=\langle \mathbf{x},\beta_{\ast}\rangle$, where the link functions $G$ is chosen as the logit function in the logistic regression, the log function in the Poisson regression, the identity function in the linear regression, etc.
If we denote the logit function, the log function, and the identity function by $G^{1}$, $G^{2}$, and $G^{3}$, respectively, we have that
\[G^{1}(t)=\log \left(\frac{t}{1-t}\right),\quad  G^{2}(t)=e^{t},\quad G^{3}(t)=t.\]

In the generalized linear model, the ground-truth parameter $\beta_{\ast}$ is estimated by the maximum likelihood estimation method, and the associated loss function can be denoted by $L(f(\mathbf{x},\beta),y)=L(\langle\mathbf{x},\beta\rangle,y)$.
If we denote the loss function in the logistic regression, the Poisson regression and the linear regression by $L^{1}$, $L^{2}$, and $L^{3}$, respectively, we have that
\[L^{1}\left(\langle \mathbf{x},\beta\rangle,y\right)=\log(1+e^{-y\langle \mathbf{x},\beta\rangle}),\]
\[L^{2}\left(\langle\mathbf{x},\beta\rangle,y\right)=e^{\langle \mathbf{x},\beta\rangle}-y\langle \mathbf{x},\beta\rangle,\]
\[L^{3}(\langle\mathbf{x},\beta\rangle,y)=\frac{1}{2}\left(\langle\mathbf{x},\beta\rangle-y\right)^2,\]
where $\beta\in B$, $B$ is a compact convex subset of $\mathbb{R}^d$, $\beta_{\ast}\in {B}^{\circ}$, $\mathbf{x}\in\Omega$, and  $\Omega$ is a convex subset of $\mathbb{R}^d$.

\subsection{Asymptotic Convergence of the WDRO Estimator}\label{example}
This subsection derives the convergence of the WDRO estimator $\beta^{DRO}_n$ in the linear regression, logistic regression, and Poisson regression.

Suppose that our choice of hyperparameters follows Assumption \ref{assume1}.
As demonstrated in Section \ref{section2.1}, we check Assumption \ref{assumeloss} and Assumption \ref{assumedistr} in the following lemmas.
\begin{lemma}\label{checklemma1}
    The loss function $L^{1}(\langle \mathbf{x},\beta\rangle,y)$ satisfies the conditions in Assumption \ref{assumeloss}.
\end{lemma}
\begin{lemma}\label{poissonassume}
If $\Omega$ is bounded, the loss function $L^{2}(\langle\mathbf{x},\beta\rangle,y)$ satisfies the conditions Assumption \ref{assumeloss}.
\end{lemma}
\begin{lemma}\label{linearassume}
The loss function $L^{3}(\langle\mathbf{x},\beta\rangle,y)$ satisfies the conditions Assumption \ref{assumeloss}.
\end{lemma}
\begin{lemma}\label{checklemma2}
    In the logistic regression, if there does not exist nonzero vector $\alpha$ such that $P_{\ast}(\alpha^{\top}\mathbf{X}=0)=1$, and $\mathbb{E}_{P_\ast}\left[\Vert\mathbf{X}\Vert_2^2\right]<\infty$, Assumption \ref{assumedistr} is satisfied.
\end{lemma}
\begin{lemma}\label{poissonassume2}
        In the Poisson regression, if there does not exist nonzero vector $\alpha$ such that $P_{\ast}(\alpha^{\top}\mathbf{X}=0)=1$, and $\mathbb{E}_{P_{\ast}}[ e^{\langle\mathbf{X},\beta_\ast\rangle}\Vert\mathbf{X}\Vert_2^2]<\infty$, Assumption \ref{assumedistr} is satisfied.
\end{lemma}
\begin{lemma}\label{linearassume2}
        In the linear regression, if there does not exist nonzero vector $\alpha$ such that $P_{\ast}(\alpha^{\top}\mathbf{X}=0)=1$, $\Var_{P_\ast}(Y\vert \mathbf{X})<\infty$, and $\mathbb{E}_{P_{\ast}}[ \Vert\mathbf{X}\Vert_2^2]<\infty$, Assumption \ref{assumedistr} is satisfied.
\end{lemma}

Lemma \ref{checklemma1}-\ref{linearassume} imply that the loss functions satisfy the conditions in Assumption \ref{assumeloss} while Lemma \ref{checklemma2}-\ref{linearassume2} show that Assumption \ref{assumedistr} can be simplified in the logistic regression, Poisson regression, and linear regression.

Equipped with Lemma \ref{checklemma1}-\ref{linearassume2}, the convergence in distribution of the WDRO estimator $\beta_n^{DRO}$ can be established due to Theorem \ref{theorem1}. 
The following three propositions give the explicit expression of the asymptotic distribution of the WDRO estimator for the logistic regression, Poisson regression, and linear regression.
\begin{proposition}[Convergence of $\beta_n^{DRO}$ in the logistic regression]\label{logistic} In the logistic regression, under Assumption \ref{assume1}, if $\Omega=\mathbb{R}^d$ and $\mathbb{E}_{P_\ast}\left[\Vert\mathbf{X}\Vert_2^2\right]<\infty$, and there does not exist nonzero vector $\alpha$ such that $P_{\ast}(\alpha^{\top}\mathbf{X}=0)=1$, the WDRO estimator $\beta_n^{DRO}$ converges in distribution:
\[\sqrt{n}(\beta_{n}^{DRO}-\beta_\ast)\Rightarrow\mathcal{N}(-C(\beta_\ast)^{-1}H(\beta_\ast),D(\beta_\ast)),\]
where
\begin{equation}\label{cov1}D(\beta_\ast)=\left(\mathbb{E}_{P_{\ast}}\left[ \frac{e^{\langle\mathbf{X},\beta_\ast\rangle}\mathbf{X}\mathbf{X}^{\top}}{\left( 1+e^{\langle\mathbf{X},\beta_{\ast}\rangle}\right)^2}\right]\right)^{-1} ,\end{equation}
and
\begin{equation}\label{H1}
C(\beta_\ast)=\mathbb{E}_{P_{\ast}}\left[ \frac{e^{\langle\mathbf{X},\beta_\ast\rangle}\mathbf{X}\mathbf{X}^{\top}}{\left( 1+e^{\langle\mathbf{X},\beta_{\ast}\rangle}\right)^2}\right],\quad
H(\beta_\ast)=\tau\sqrt{\mathbb{E}_{P_{\ast}}\left[\frac{e^{\langle\mathbf{X},\beta_{\ast}\rangle}}{\left(1+e^{\langle\mathbf{X},\beta_{\ast}\rangle}\right)^2}\right]}\frac{\beta_\ast}{\Vert\beta_\ast\Vert_2}.\end{equation}
\end{proposition}
\begin{proposition}[Convergence of $\beta_n^{DRO}$ in the Poisson regression]\label{poissontheorem}
In the Poission regression, under Assumption \ref{assume1}, if $\Omega$ is compact and can be expressed as $\Omega=\{\mathbf{x}\in\mathbb{R}^d: A\mathbf{x}\leq b\}$, where $A$ is an $l\times d$ matrix with linearly independent rows and $b\in\mathbb{R}^l$, $\mathbb{E}_{P_{\ast}}[ \Vert\mathbf{X}\Vert_2^2e^{\langle\mathbf{X},\beta_\ast\rangle}]<\infty$, there does not exist nonzero vector $\alpha$ such that $P_{\ast}(\alpha^{\top}\mathbf{X}=0)=1$, and $\mathbf{X}$ has a probability density which is absolutely continuous w.r.t. Lebesgue measure, the WDRO estimator $\beta_n^{DRO}$ converges in distribution:
\[\sqrt{n}(\beta_{n}^{DRO}-\beta_\ast)\Rightarrow\mathcal{N}(-C(\beta_\ast)^{-1}H(\beta_\ast),D(\beta_\ast)),\]
where
\begin{equation}\label{cov2}D(\beta_\ast)= \left(\mathbb{E}_{P_{\ast}}\left[ e^{\langle \mathbf{X},\beta_{\ast}\rangle}\mathbf{X}\mathbf{X}^{\top}\right]\right)^{-1},\end{equation}
and
\begin{equation}\label{H2}
C(\beta_\ast)=\mathbb{E}_{P_{\ast}}\left[ e^{\langle \mathbf{X},\beta_{\ast}\rangle}\mathbf{X}\mathbf{X}^{\top}\right],\quad
H(\beta_{\ast})=\tau\sqrt{\mathbb{E}_{P_\ast}[e^{\langle \mathbf{X},\beta_{\ast}\rangle}]}\frac{\beta_{\ast}}{\Vert\beta_\ast\Vert_2}.\end{equation}
\end{proposition}
\begin{proposition}[Convergence of $\beta_n^{DRO}$ in the linear regression]
\label{lineartheorem}
In the linear regression, under Assumption \ref{assume1}, if $\Omega=\mathbb{R}^d$, and $\mathbb{E}_{P_{\ast}}[ \Vert\mathbf{X}\Vert_2^2]<\infty$, and there does not exist nonzero vector $\alpha$ such that $P_{\ast}(\alpha^{\top}\mathbf{X}=0)=1$, the WDRO estimator $\beta_n^{DRO}$ converges in distribution:
\[\sqrt{n}(\beta_{n}^{DRO}-\beta_\ast)\Rightarrow\mathcal{N}(-C^{-1}H(\beta_\ast),D),\]
where
\begin{equation}\label{cov3}D=\sigma^2\left(\mathbb{E}_{P_{\ast}}\left[ \mathbf{X}\mathbf{X}^{\top}\right]\right)^{-1},\end{equation}
\begin{equation}\label{H3}C=\mathbb{E}_{P_{\ast}}\left[ \mathbf{X}\mathbf{X}^{\top}\right],\quad
H(\beta_{\ast})=\tau\sigma\frac{\beta_{\ast}}{\Vert\beta_\ast\Vert_2},\end{equation}
 and $\Var_{P_\ast}(Y|\mathbf{X})=\sigma^2, \sigma>0$
\end{proposition}

We could obtain the associated adjusted WDRO estimators based on the convergence results derived in Proposition \ref{logistic}-\ref{lineartheorem}, and the details will be clarified in the next subsection.

Also, the proofs of Proposition \ref{logistic}-\ref{lineartheorem} are relegated to Appendix \ref{appendix}. The proofs show that the conditions in Proposition \ref{sufficient2} are satisfied, which enables us to simplify the function $H$, seeing \eqref{H1}, \eqref{H2} and \eqref{H3}.

\subsection{Adjusted WDRO Estimator in the Generalized Linear Model}
This subsection gives the formulations of the adjusted WDRO estimator for logistic regression, Poisson regression, and linear regression by plugging the expressions of the function $C$ and $H$ in \eqref{H1}, \eqref{H2} and \eqref{H3} into the definition of the adjusted WDRO estimator \eqref{adrodefinition}.
\begin{definition}\label{defexample}
Under assumptions in Proposition \ref{logistic}-\ref{lineartheorem}, for the nonzero WDRO estimator $\beta_n^{DRO}$, we define the adjusted WDRO estimator $\beta_n^{ADRO}$ as follows, 
\begin{equation}\label{computeadrolog} \beta_n^{ADRO}= \beta_n^{DRO}+\frac{\tau}{\sqrt{n}}\sqrt{\mathbb{E}_{\mathbb{P}_n}\left[\frac{e^{\langle\mathbf{X},\beta_n^{DRO}\rangle}}{\left(1+e^{\langle\mathbf{X},\beta_n^{DRO}\rangle}\right)^2}\right]}\left(\mathbb{E}_{\mathbb{P}_n}\left[\frac{e^{\langle\mathbf{X},\beta_n^{DRO}\rangle}\mathbf{X}\mathbf{X}^{\top}}{\left(1+e^{\langle\mathbf{X},\beta_n^{DRO}\rangle}\right)^2}\right]\right)^{-1}\frac{\beta_n^{DRO}}{\Vert\beta_n^{DRO}\Vert_2}, \end{equation}
\[\beta_n^{ADRO}=\beta_n^{DRO}+\frac{\tau}{\sqrt{n}}\sqrt{\mathbb{E}_{\mathbb{P}_n}[e^{\langle \mathbf{X},\beta_n^{DRO}\rangle}]}\left(\mathbb{E}_{\mathbb{P}_n}\left[e^{\langle\mathbf{X},\beta_n^{DRO}\rangle} \mathbf{X}\mathbf{X}^{\top}  
\right]\right)^{-1}\frac{\beta_n^{DRO}}{\Vert\beta_n^{DRO}\Vert_2},\]
\begin{equation}\label{adjustdrolinear}\beta_n^{ADRO}=\beta_n^{DRO}+\frac{\tau\sigma}{\sqrt{n}} \left(\mathbb{E}_{\mathbb{P}_n}\left[ \mathbf{X}\mathbf{X}^\top\right]\right)^{-1} \frac{\beta_n^{DRO}}{\Vert\beta_n^{DRO}\Vert_2},\end{equation}
for the logistic regression, Poisson regression, and linear regression, respectively.
\end{definition}

As we discussed in Proposition \ref{conditions2}, one could check that the adjusted WDRO estimators defined in Definition \ref{defexample} are well-defined.
Then, it is easy to check that the conditions in Theorem \ref{theorem2}, i.e., the smoothness of the loss function,  hold for the logistic regression, Poisson regression, and linear regression, indicating the proposed adjustment technique could de-bias the associated adjusted WDRO estimators successfully in the logistic regression, Poisson regression, and linear regression.
We conclude this result in the following proposition.
\begin{proposition}\label{unbiasedglm}
    For the adjusted WDRO estimator $\beta_n^{ADRO}$ defined in Definition \ref{defexample}, we have the following 
    \[\sqrt{n}\left(\beta_n^{ADRO}-\beta_\ast\right)\Rightarrow \mathcal{N}(0,D(\beta_\ast)),\]
    where $D(\beta_\ast)$ is defined by \eqref{cov1}, \eqref{cov2}, and \eqref{cov3} in the logistic regression, Poisson regression, and linear regression, respectively.
\end{proposition}
\section{Numerical Experiments}\label{exp}
In this section, we investigate the empirical performance of the adjusted WDRO estimator $\beta_n^{ADRO}$, compared with the classic WDRO estimator $\beta_n^{DRO}$.

\subsection{Experiment Setting}
The WDRO algorithmic framework of the logistic regression model and linear regression model with quadratic loss has been established in \cite{blanchet2022optimal}. 
Therefore, the adjusted WDRO estimators in the logistic regression model and the
linear regression model are implemented as examples to evaluate the practical performance of our adjustment technique.
\subsubsection{Logistic Regression}
Suppose $\mathbf{X}$ follows 2-dimensional standard normal distribution, and the response variable $Y$ follows the Bernoulli distribution, where $P_\ast(Y=1|\mathbf{X}=\mathbf{x})=1/(1+e^{-\langle \mathbf{x},\beta_{\ast}\rangle})$ and $\beta_\ast=(1/\sqrt{17},4/\sqrt{17})$. 
Data is generated $5$ times for each sample size $n\in\{500,700,1000,1500,$ $1800,2000\}$. 
The WDRO estimator $\beta_n^{DRO}$ is computed by the iterative algorithm in \cite{blanchet2022optimal}.
The adjusted WDRO estimator $\beta_n^{ADRO}$ is computed via equation \eqref{computeadrolog}.
Per the iterative algorithm, we set the learning rate as $0.3$ and the maximum number of iterations as $50000$, respectively.
Moreover, since the value of $\tau$, which is the coefficient in the Wasserstein radius $\rho_n=\tau/\sqrt{n}$, should be determined,  we let $\tau\in\{1.5,2,2.5,3\}$. 

\subsubsection{Linear regression}
Assume the feature variable $\mathbf{X}$ follows the 2-dimensional standard normal distribution, and the response variable $Y$ follows the normal distribution, where $Y|\mathbf{X}=\mathbf{x}\sim \mathcal{N}(\langle\mathbf{x},\beta_\ast\rangle,\sigma)$, $\beta_\ast=(3/\sqrt{10},-1/\sqrt{10})$. We set $\sigma=0.1$.
Data is generated $5$ times for each sample size $n\in\{500,700,1000,1500,1800,2000\}$. 
The WDRO estimator $\beta_n^{DRO}$ is computed by the iterative algorithm in \cite{blanchet2022optimal}.
The adjusted WDRO estimator $\beta_n^{ADRO}$ is computed via equation \eqref{adjustdrolinear}.
Per the iterative algorithm, we set the learning rate as $0.01$  and the maximum number of iterations as $50000$, respectively.
Then, we set the value of $\tau$ as $\tau\in\{1.5,2,2.5,3\}$.
\subsection{Experiment Results}
\begin{figure}
    \centering
    \begin{minipage}{0.4\textwidth}
        \centering
        \includegraphics[width=\textwidth]{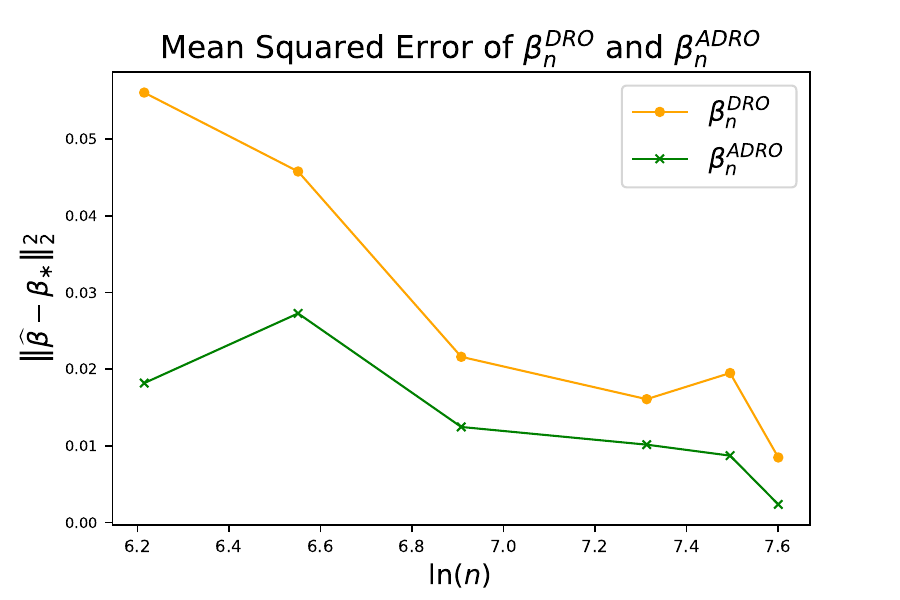} 
    \end{minipage}\hspace{0.1in}
    \begin{minipage}{0.4\textwidth}
        \centering
        \includegraphics[width=\textwidth]{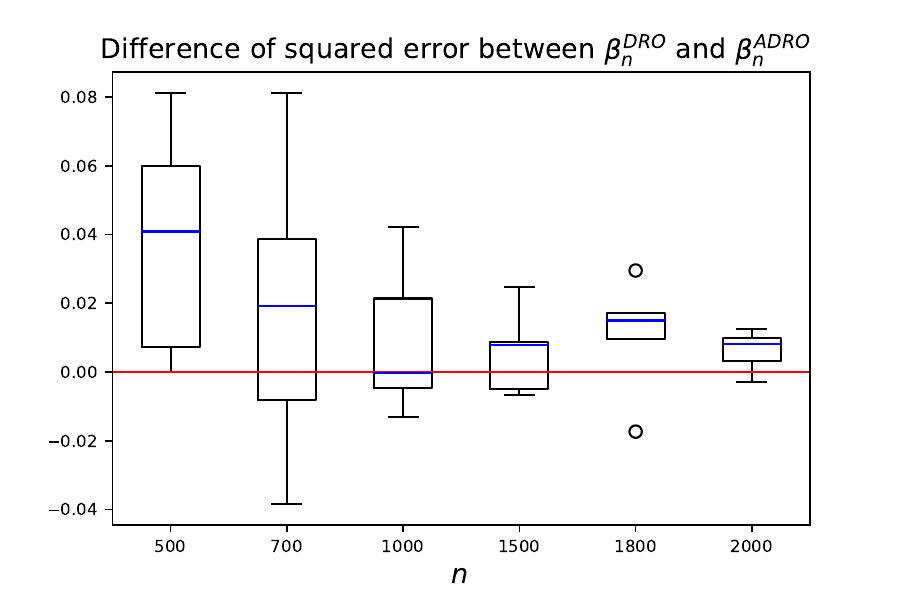} 
    \end{minipage}\\
    \begin{minipage}{0.4\textwidth}
        \centering
        \includegraphics[width=\textwidth]{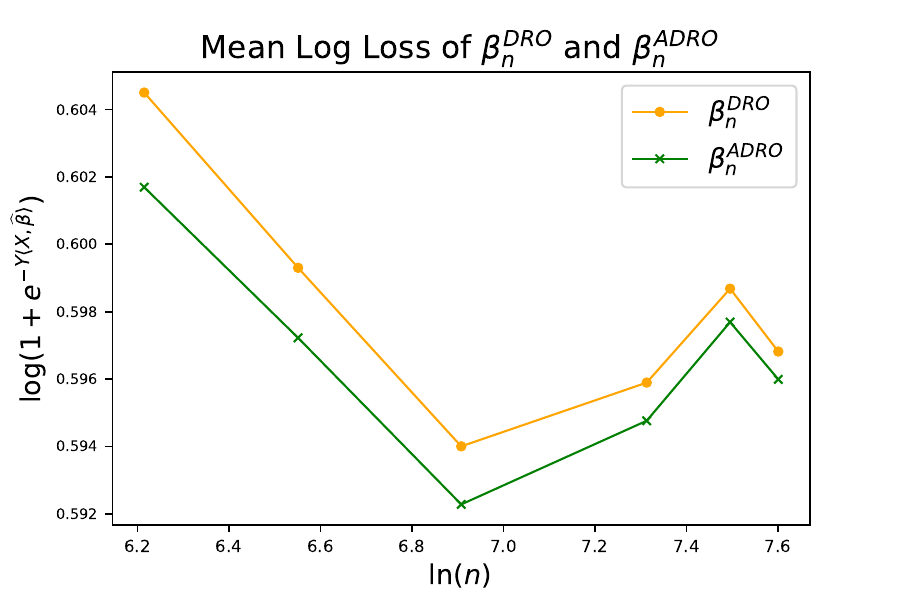} 
    \end{minipage}\hspace{0.1in}
    \begin{minipage}{0.4\textwidth}
        \centering
        \includegraphics[width=\textwidth]{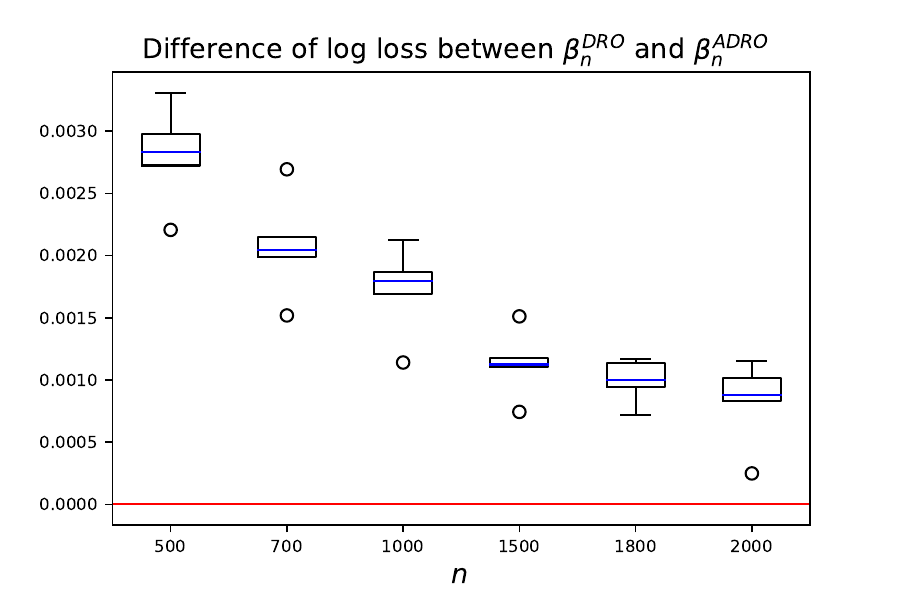} 
    \end{minipage}
    \caption{Squared error and log loss plots of the logistic regression, $\tau=1.5$.}
    \label{logistic1}
\end{figure}

\begin{figure}
    \centering
    \begin{minipage}{0.4\textwidth}
        \centering
        \includegraphics[width=\textwidth]{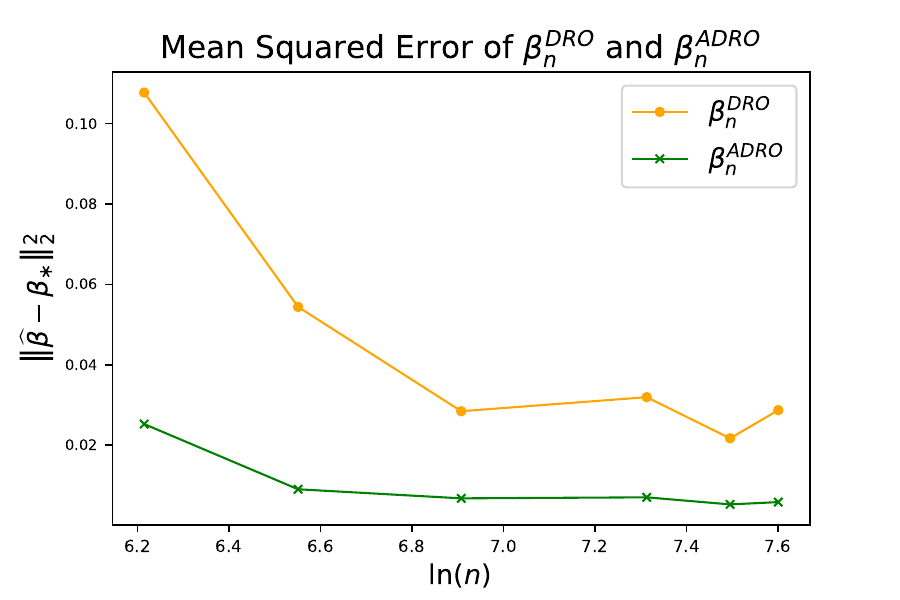} 
    \end{minipage}\hspace{0.1in}
    \begin{minipage}{0.4\textwidth}
        \centering
        \includegraphics[width=\textwidth]{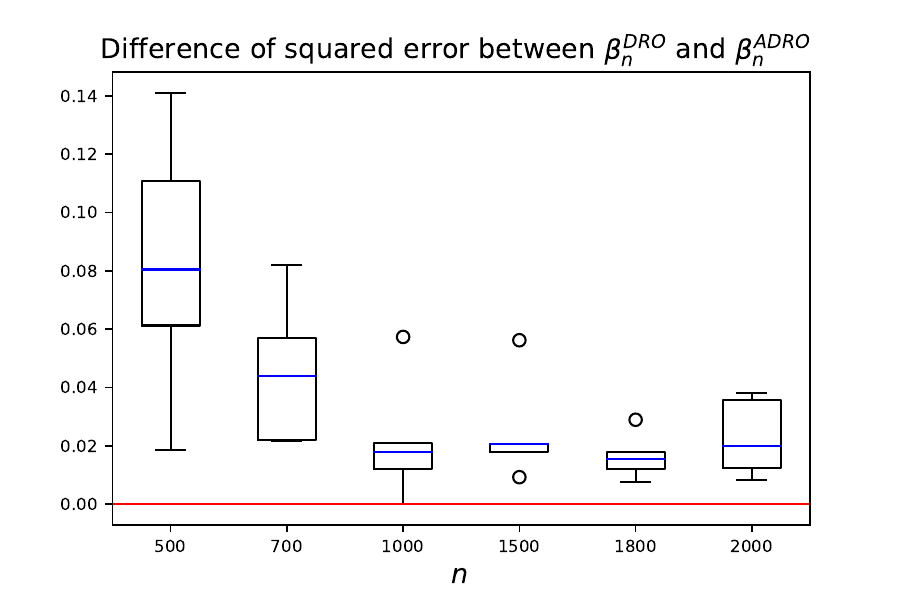} 
    \end{minipage}\\
    \begin{minipage}{0.4\textwidth}
        \centering
        \includegraphics[width=\textwidth]{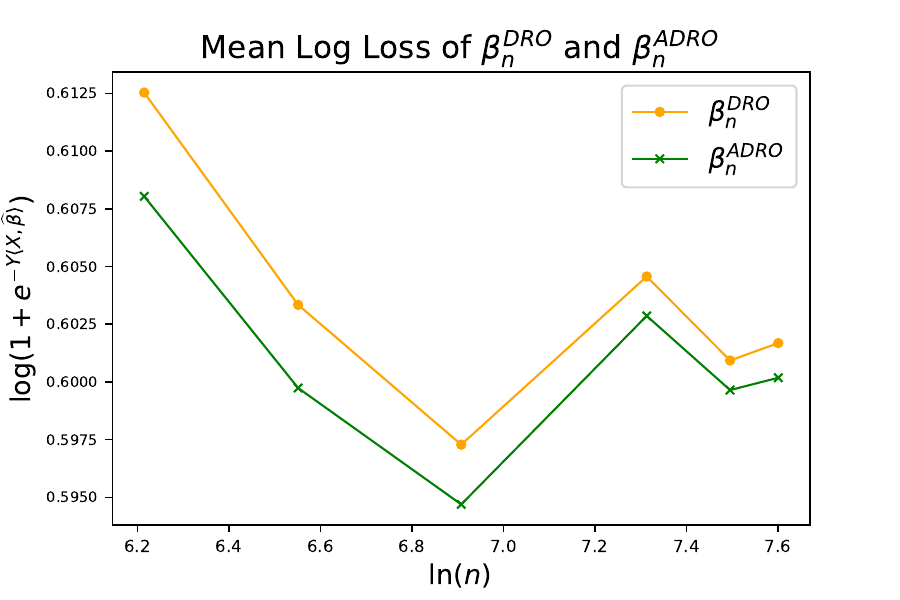} 
    \end{minipage}\hspace{0.1in}
    \begin{minipage}{0.4\textwidth}
        \centering
        \includegraphics[width=\textwidth]{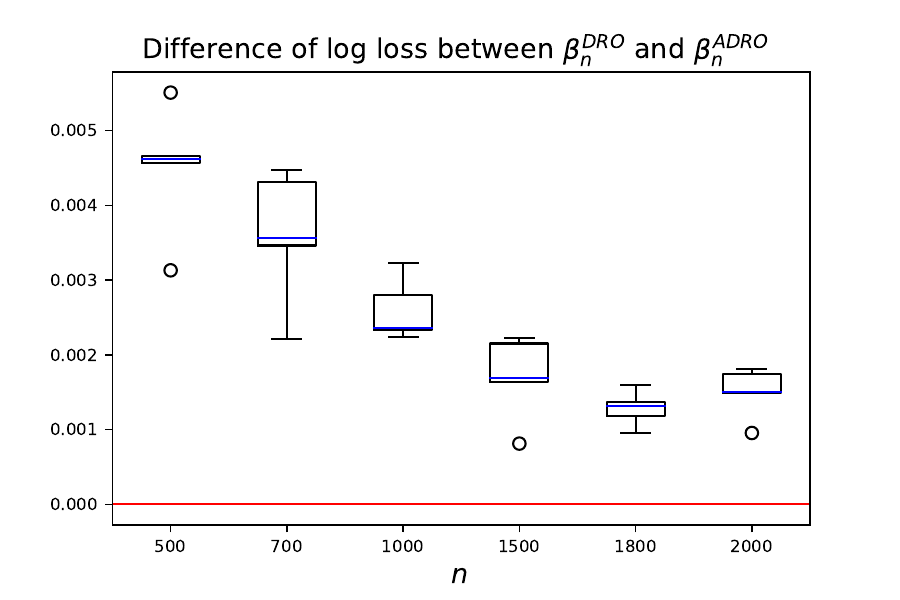} 
    \end{minipage}
    \caption{Squared error and log loss plots of the logistic regression, $\tau=2$.}
    \label{logistic2}
\end{figure}

\begin{figure}
    \centering
    \begin{minipage}{0.4\textwidth}
        \centering
        \includegraphics[width=\textwidth]{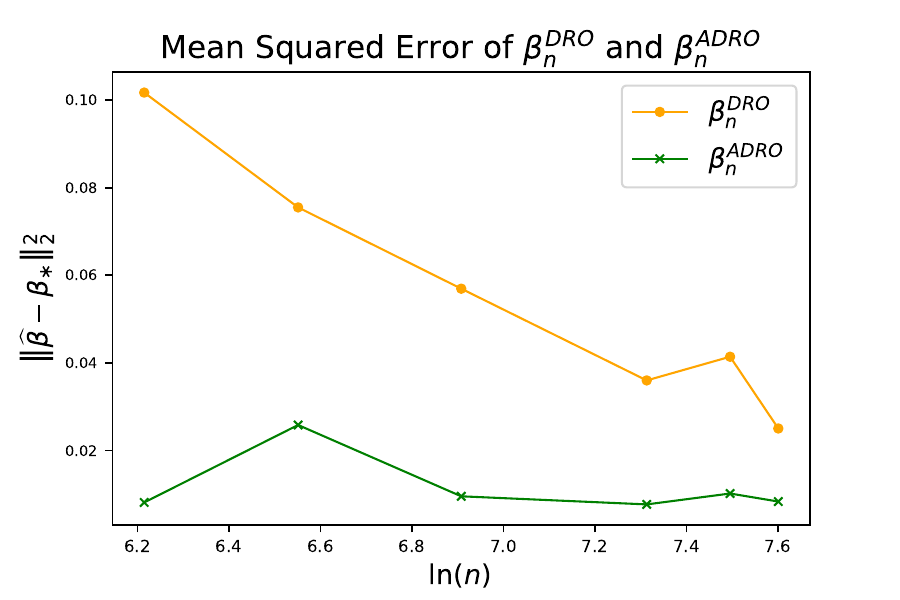} 
    \end{minipage}\hspace{0.1in}
    \begin{minipage}{0.4\textwidth}
        \centering
        \includegraphics[width=\textwidth]{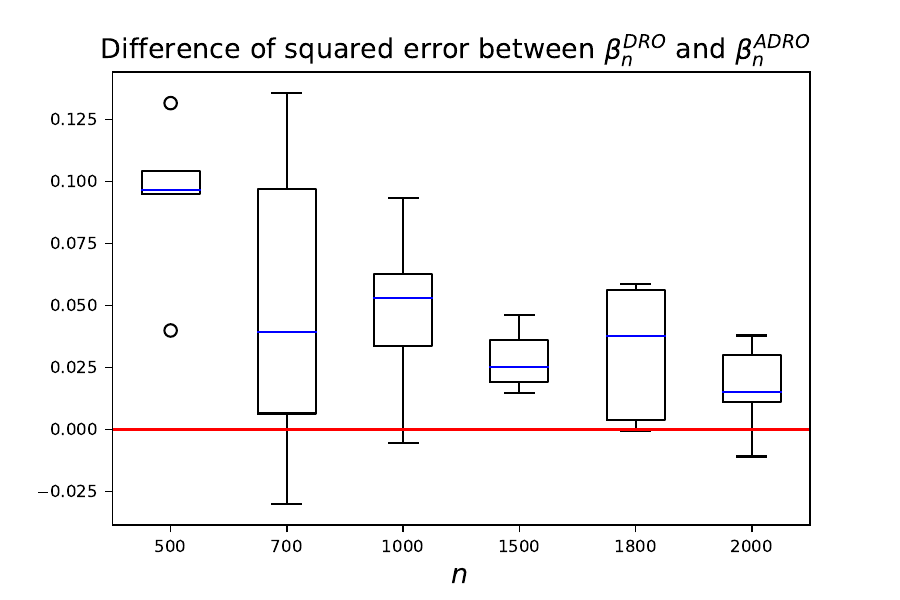} 
    \end{minipage}\\
    \begin{minipage}{0.4\textwidth}
        \centering
        \includegraphics[width=\textwidth]{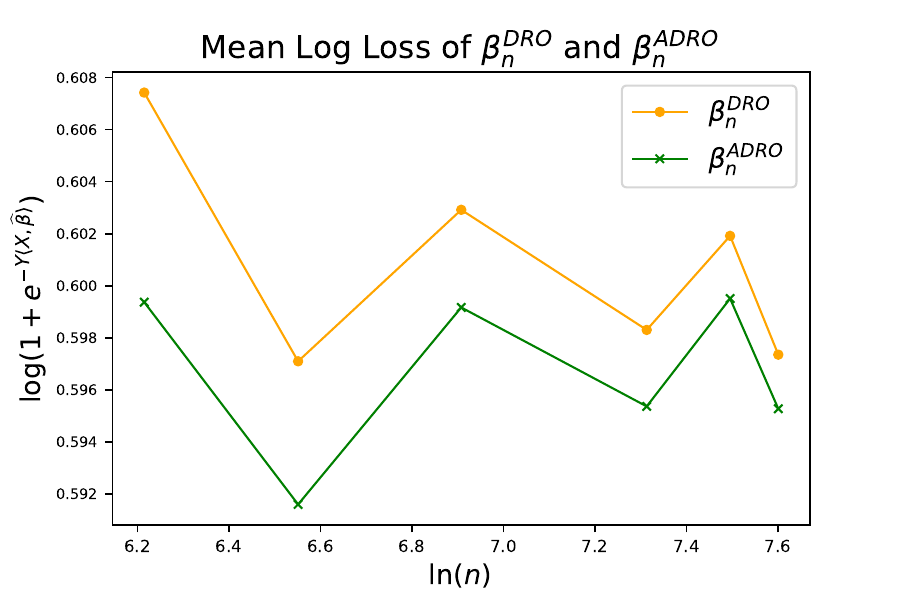} 
    \end{minipage}\hspace{0.1in}
    \begin{minipage}{0.4\textwidth}
        \centering
        \includegraphics[width=\textwidth]{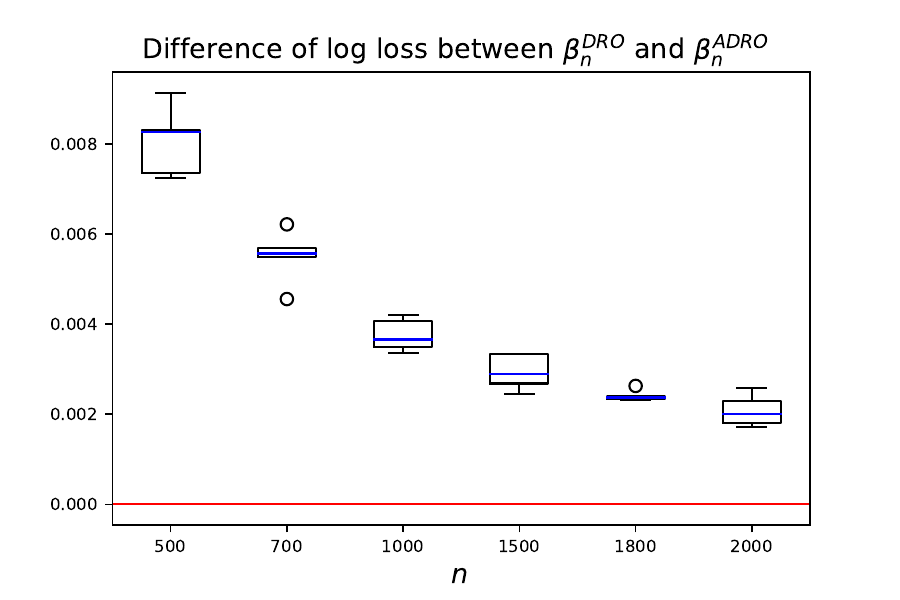} 
    \end{minipage}
    \caption{Squared error and log loss plots of the logistic regression, $\tau=2.5$.}
    \label{logistic3}
\end{figure}

\begin{figure}
    \centering
    \begin{minipage}{0.4\textwidth}
        \centering
        \includegraphics[width=\textwidth]{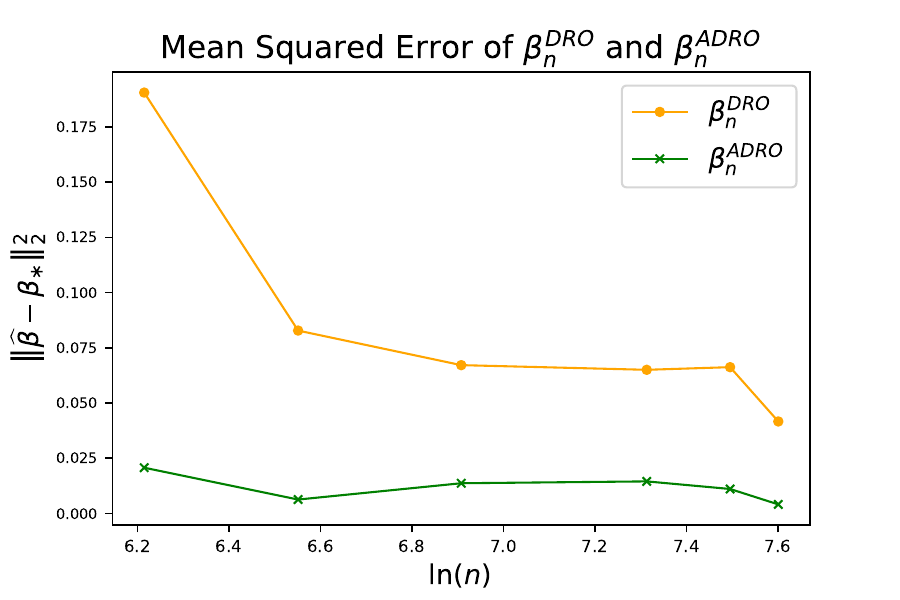} 
    \end{minipage}\hspace{0.1in}
    \begin{minipage}{0.4\textwidth}
        \centering
        \includegraphics[width=\textwidth]{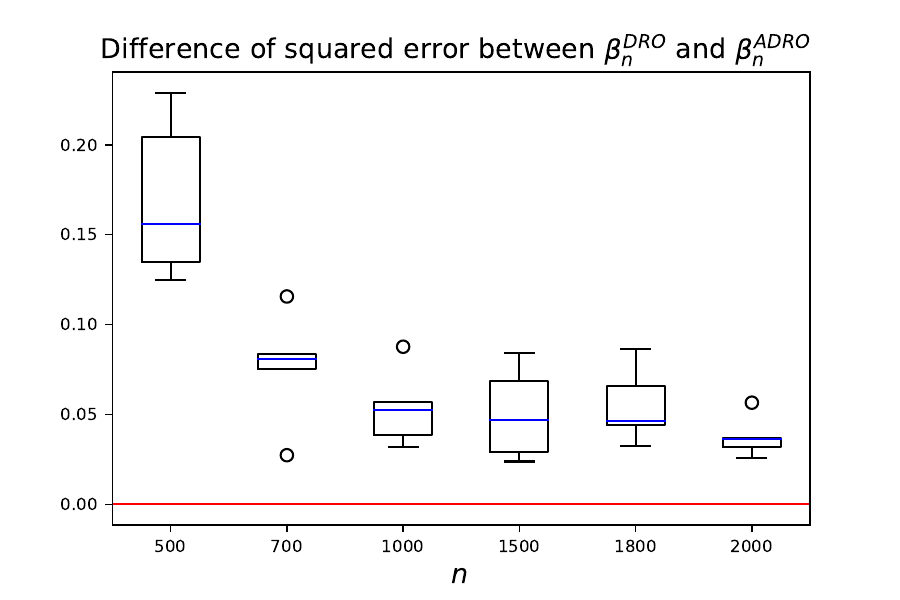} 
    \end{minipage}\\
    \begin{minipage}{0.4\textwidth}
        \centering
        \includegraphics[width=\textwidth]{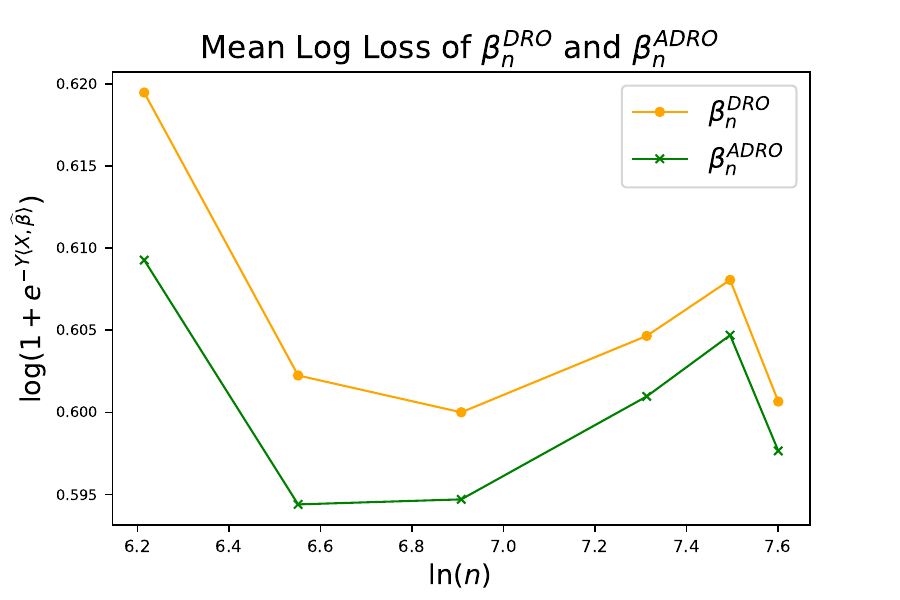} 
    \end{minipage}\hspace{0.1in}
    \begin{minipage}{0.4\textwidth}
        \centering
        \includegraphics[width=\textwidth]{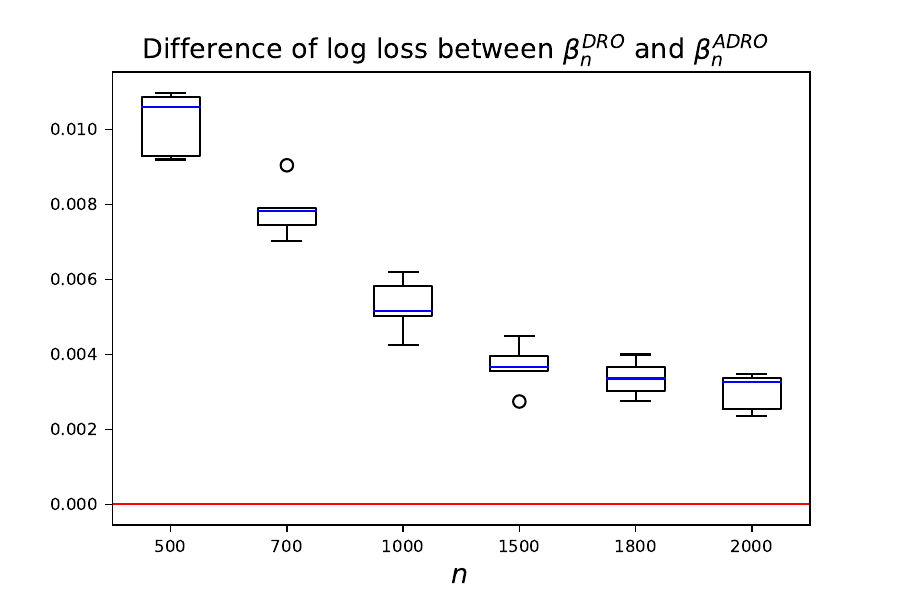} 
    \end{minipage}
    \caption{Squared error and log loss plots of the logistic regression, $\tau=3$.}
    \label{logistic4}
\end{figure}

\begin{figure}
    \centering
    \begin{minipage}{0.4\textwidth}
        \centering
        \includegraphics[width=\textwidth]{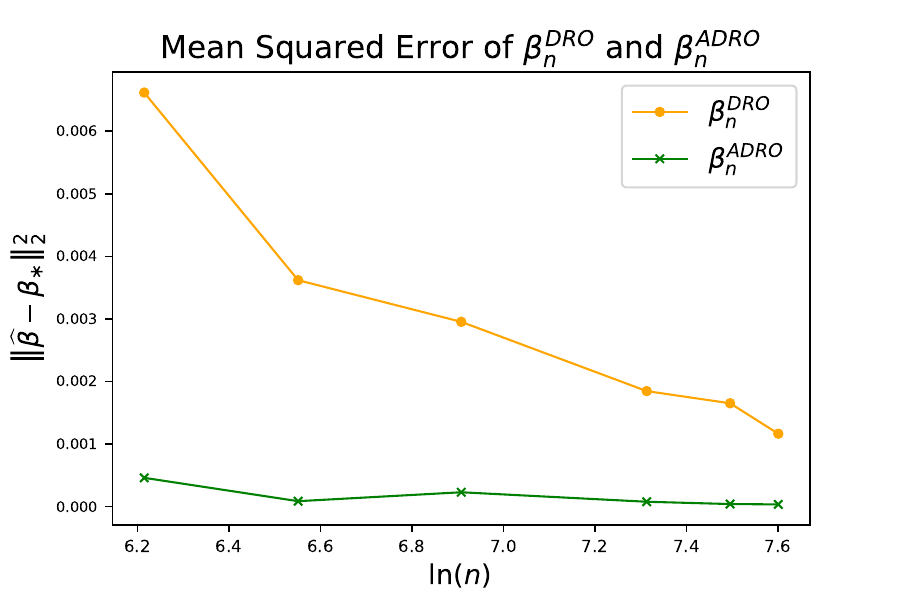} 
    \end{minipage}\hspace{0.1in}
    \begin{minipage}{0.4\textwidth}
        \centering
        \includegraphics[width=\textwidth]{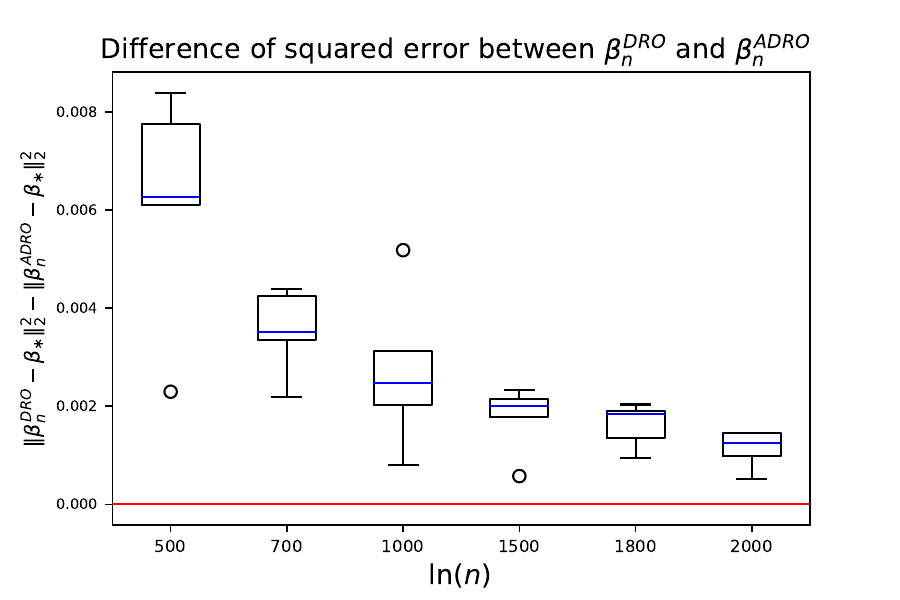} 
    \end{minipage}\\
    \begin{minipage}{0.4\textwidth}
        \centering
        \includegraphics[width=\textwidth]{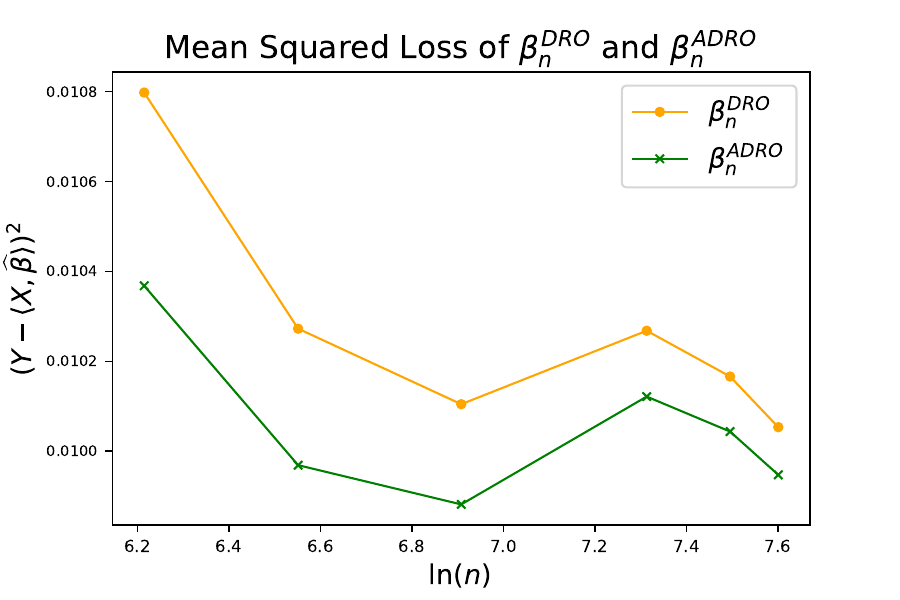} 
    \end{minipage}\hspace{0.1in}
    \begin{minipage}{0.4\textwidth}
        \centering
        \includegraphics[width=\textwidth]{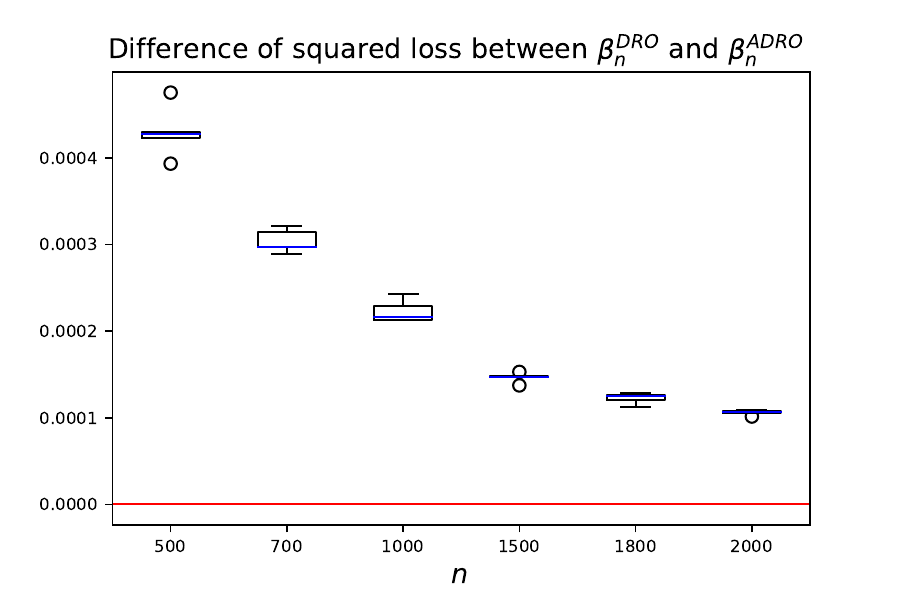} 
    \end{minipage}
    \caption{Squared error and squared loss plots of the linear regression, $\tau=1.5$.}
    \label{linear1}
\end{figure}

\begin{figure}
    \centering
    \begin{minipage}{0.4\textwidth}
        \centering
        \includegraphics[width=\textwidth]{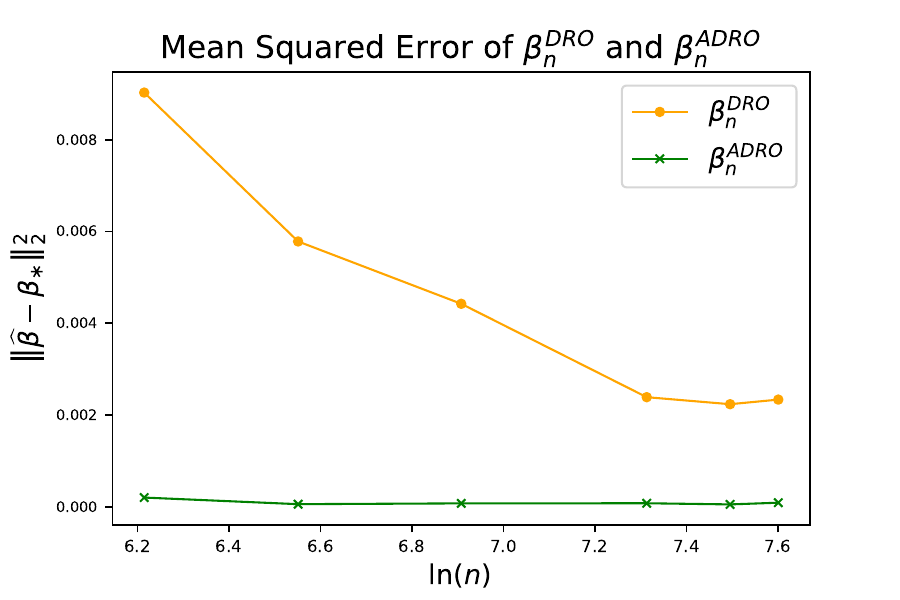} 
    \end{minipage}\hspace{0.1in}
    \begin{minipage}{0.4\textwidth}
        \centering
        \includegraphics[width=\textwidth]{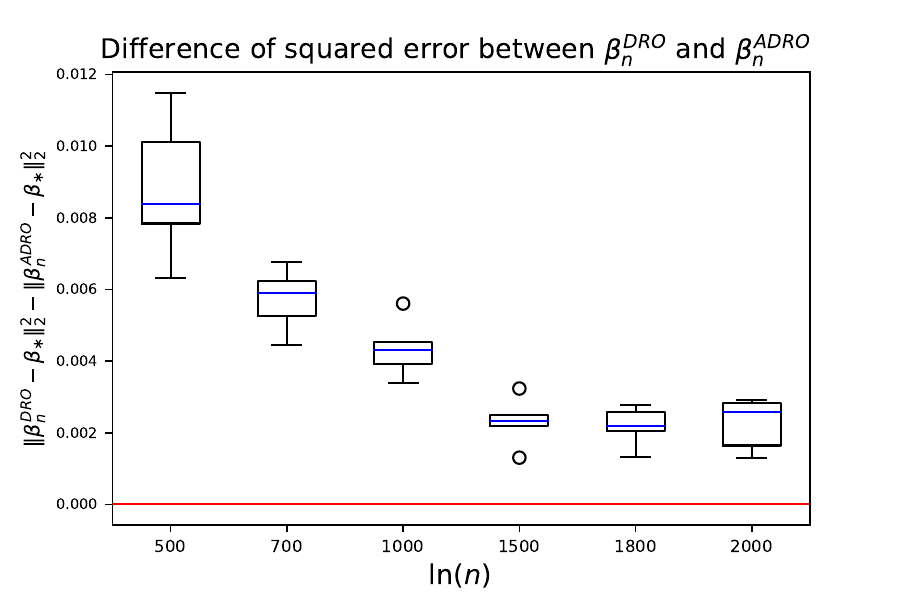} 
    \end{minipage}\\
    \begin{minipage}{0.4\textwidth}
        \centering
        \includegraphics[width=\textwidth]{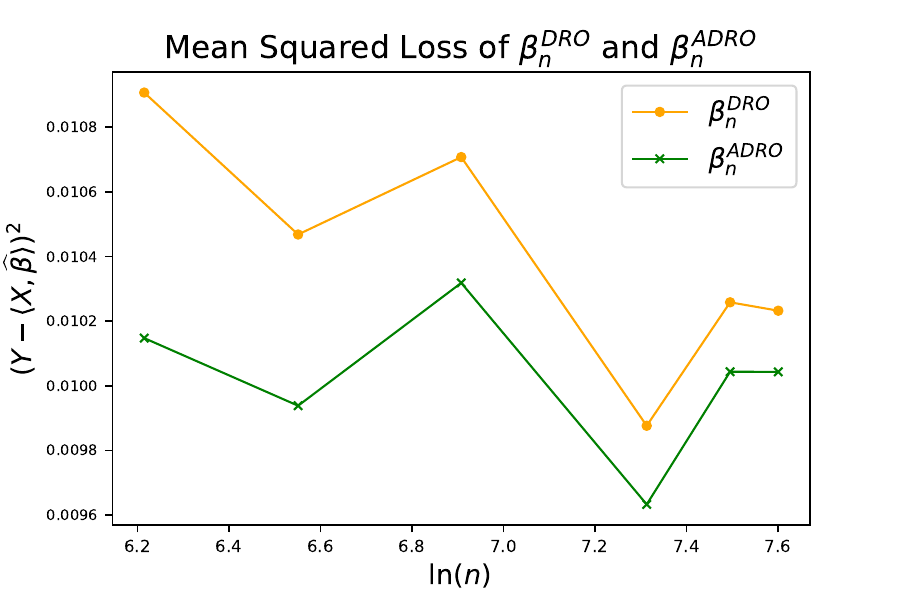} 
    \end{minipage}\hspace{0.1in}
    \begin{minipage}{0.4\textwidth}
        \centering
        \includegraphics[width=\textwidth]{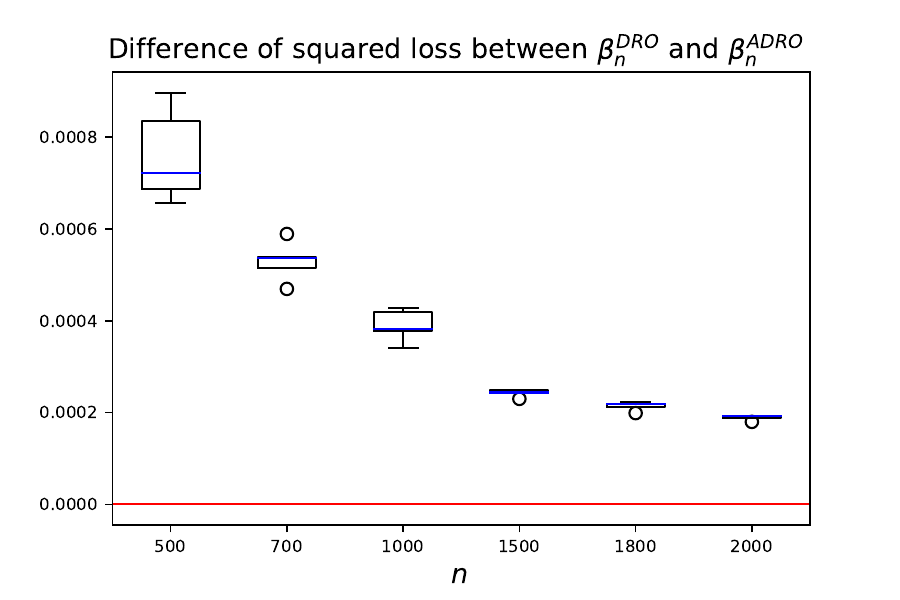} 
    \end{minipage}
    \caption{Squared error and squared loss plots of the linear regression, $\tau=2$.}
    \label{linear2}
\end{figure}

\begin{figure}
    \centering
    \begin{minipage}{0.4\textwidth}
        \centering
        \includegraphics[width=\textwidth]{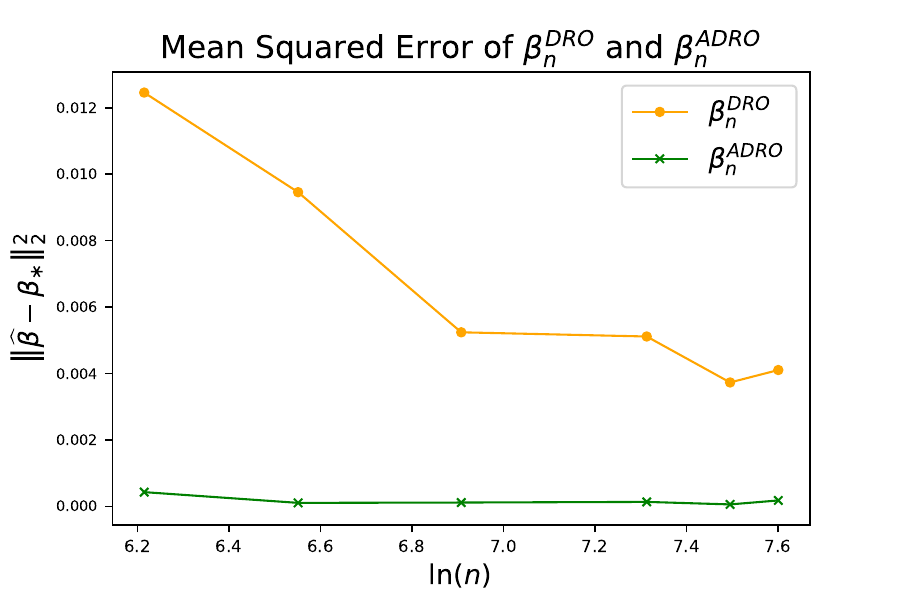} 
    \end{minipage}\hspace{0.1in}
    \begin{minipage}{0.4\textwidth}
        \centering
        \includegraphics[width=\textwidth]{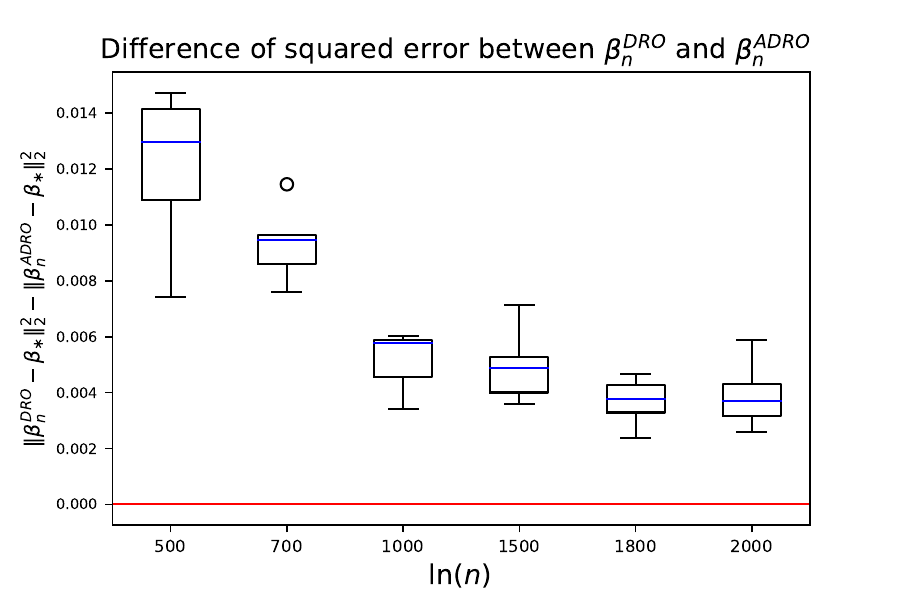} 
    \end{minipage}\\
    \begin{minipage}{0.4\textwidth}
        \centering
        \includegraphics[width=\textwidth]{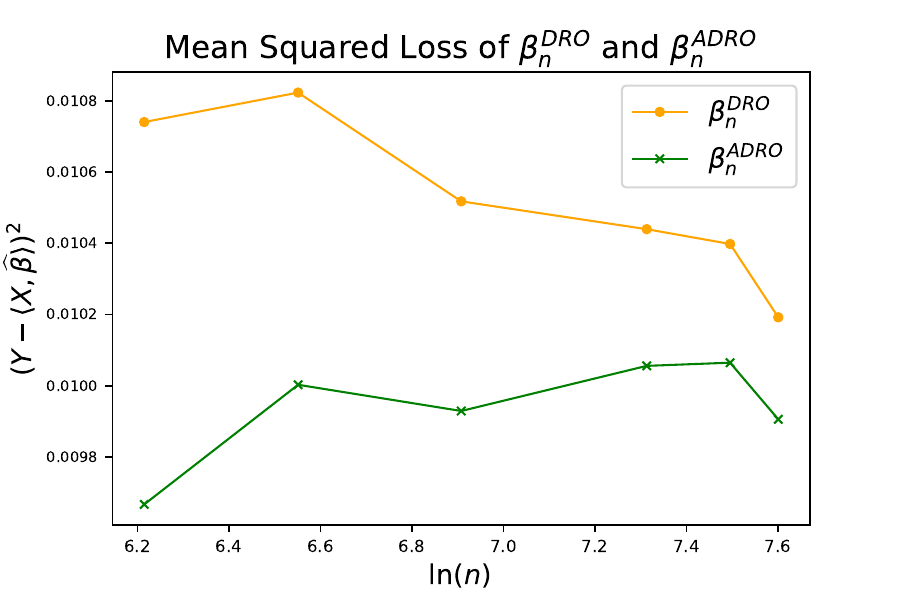} 
    \end{minipage}\hspace{0.1in}
    \begin{minipage}{0.4\textwidth}
        \centering
        \includegraphics[width=\textwidth]{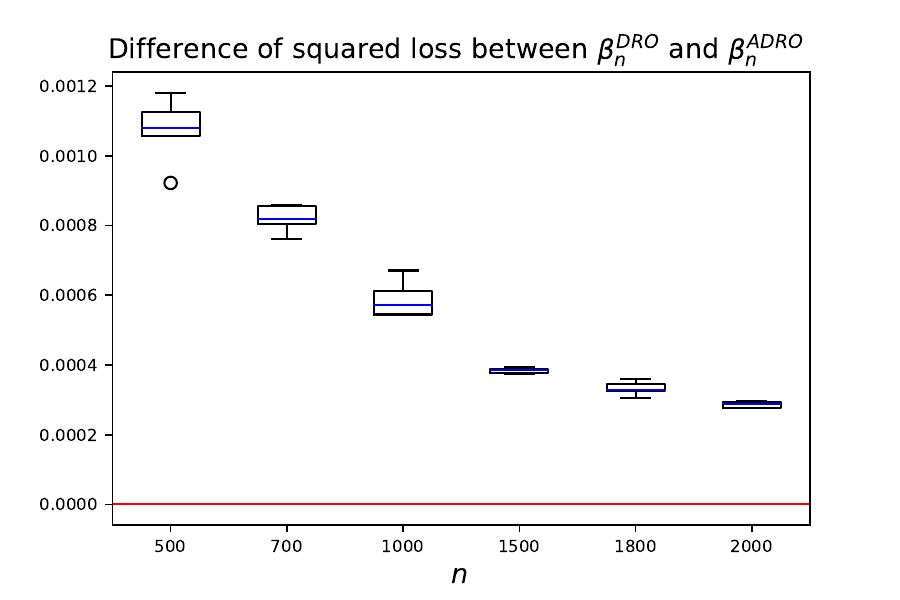} 
    \end{minipage}
    \caption{Squared error and squared loss plots of the linear regression, $\tau=2.5$.}
    \label{linear3}
\end{figure}

\begin{figure}
    \centering
    \begin{minipage}{0.4\textwidth}
        \centering
        \includegraphics[width=\textwidth]{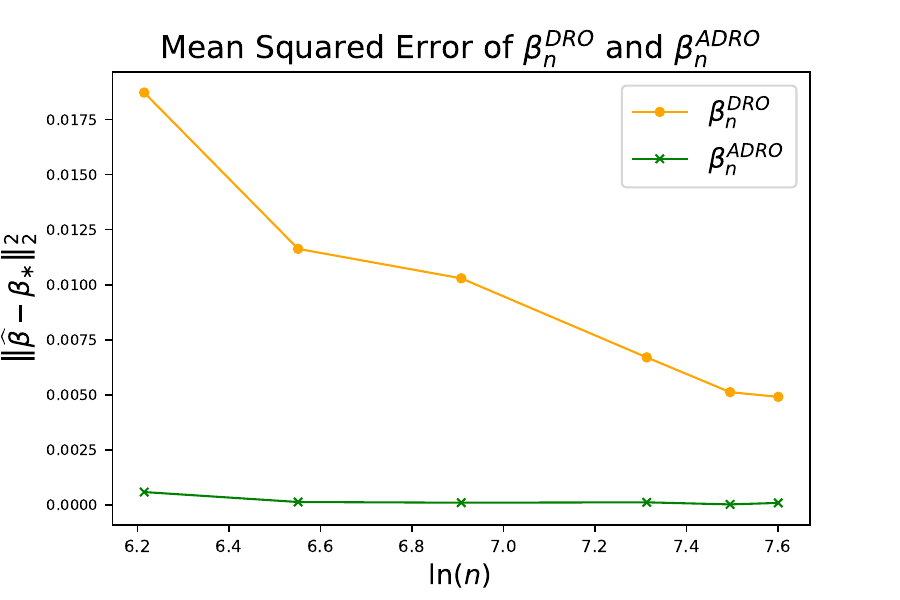} 
    \end{minipage}\hspace{0.1in}
    \begin{minipage}{0.4\textwidth}
        \centering
        \includegraphics[width=\textwidth]{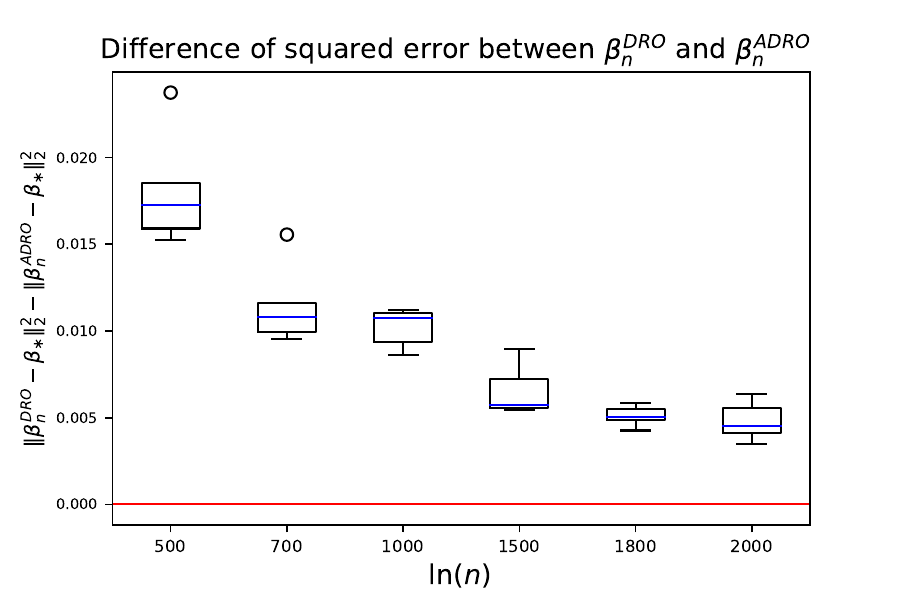} 
    \end{minipage}\\
    \begin{minipage}{0.4\textwidth}
        \centering
        \includegraphics[width=\textwidth]{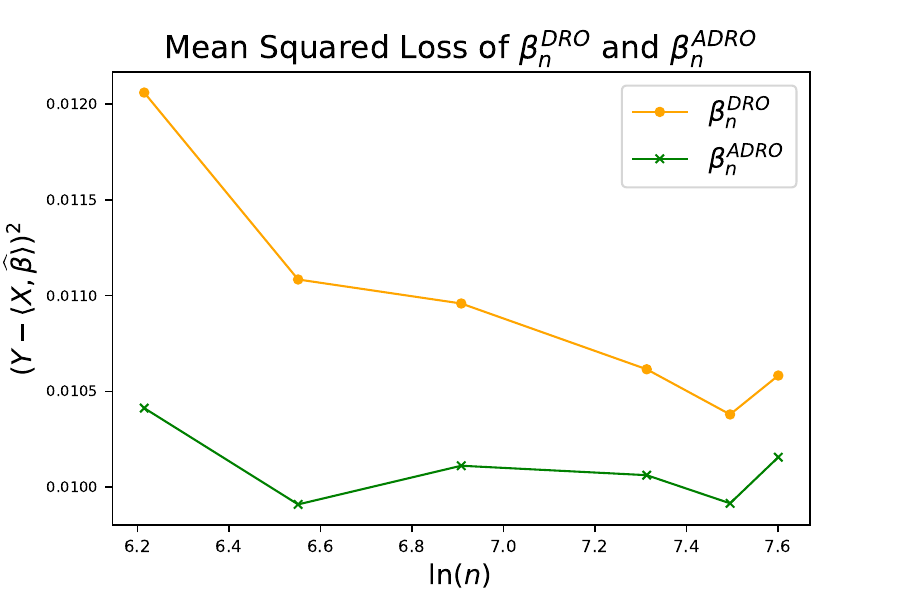} 
    \end{minipage}\hspace{0.1in}
    \begin{minipage}{0.4\textwidth}
        \centering
        \includegraphics[width=\textwidth]{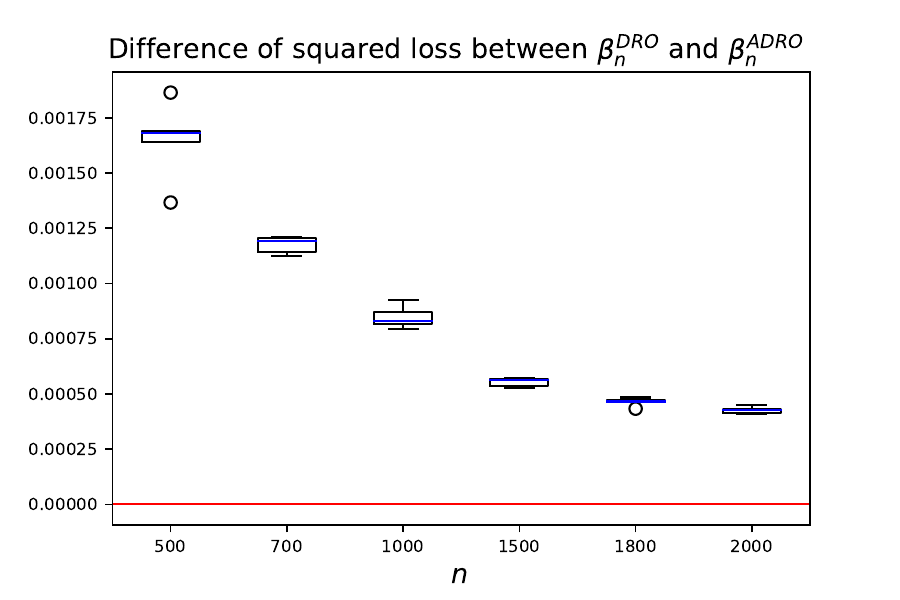} 
    \end{minipage}
    \caption{Squared error and squared  loss plots of the linear regression, $\tau=3$.}
    \label{linear4}
\end{figure}

The experimental results of the logistic regression are reported in Figure \ref{logistic1}-\ref{logistic4}, and the results of the linear regression are reported in Figure \ref{linear1}-\ref{linear4}.

The estimation accuracy of the estimators is evaluated by the squared error. 
The squared error of the estimator $\widehat{\beta}$ is defined by $\Vert \widehat{\beta}-\beta_\ast\Vert_2^2$.
We plot the mean squared error of $\beta_n^{DRO}$ and $\beta_n^{ADRO}$ versus the logarithm of the sample size $n$, respectively. 
From the figures, we observe that the line of mean squared error of $\beta_n^{DRO}$ is always above that of $\beta_n^{ADRO}$, illustrating that the proposed adjusted estimator has a smaller mean squared error. 
Recall that the adjusted WDRO estimator has a better asymptotic mean squared error in theory, while our empirical results show that the proposed estimator outperforms even when the sample size is finite.
Moreover, we compute the difference of the squared error between $\beta_n^{DRO}$ and $\beta_n^{ADRO}$ for each run.
This quantity helps evaluate the improvement achieved by the adjustment technique for each run. 
To visualize the improvement, we plot the boxplots for each sample size and each value of $\tau$. 
The figures show that most parts of the boxplots are located above $y=0$ in the logistic regression, and all of the boxplots are located above $y=0$ in the linear regression.
These observations indicate that the adjustment technique can generate a more accurate estimator for the ground-truth parameter $\beta_\ast$.

In addition to the squared error, we investigate the loss, i.e., the log-likelihood, of the estimators in the linear regression and the logistic regression. 
Similar to how we analyze the squared error, we plot the mean loss and the case-wise loss improvement. 
The figures show that the adjustment technique could help reduce the loss.

Overall, the adjusted WDRO estimator has better empirical performance than the classic WDRO estimator. When people plan to estimate parameters in statistical learning under the WDRO framework, the proposed adjusted estimator can be considered.

\section{Discussion}
This paper improves the performance of the WDRO estimator through the lens of the statistical asymptotics of the WDRO estimator.
To the best of our knowledge, we are the first to propose transformations to de-bias the WDRO estimator asymptotically.
The proposed adjusted WDRO estimator is asymptotically unbiased with a smaller asymptotic mean squared error. 
In addition, the adjusted WDRO estimator is easy to compute as long as the classic WDRO estimator is known.
Also, we observe the superior empirical performance of the adjusted WDRO estimator over the classic WDRO estimator.

Notably, we carefully clarify and check the corresponding assumptions in the development of our theory and methodology, providing a rigorous scheme for applying and generalizing our adjustment technique.
\acks{The authors would like to thank the Action Editor and anonymous reviewers for their detailed and constructive comments, which helped greatly enhance the quality and presentation of the manuscript. The authors are partially sponsored by NSF grants CCF-1740776, DMS 2015363, and IIS-2229876. They are also partially supported by the A. Russell Chandler III Professorship at Georgia Tech.}

\newpage
\appendix
\section{Proof}\label{appendix}
\subsection{Proof of Proposition \ref{uniquetransformation}}
\begin{proof}
Due to the sequential delta method, seeing Theorem \ref{extendeddeltamethod}, we have that 
\begin{equation*}\sqrt{n}\left(\phi_n(\beta_n)-\phi_n(\beta_\ast)\right)\Rightarrow\mathcal{N}(\phi^{\prime}(\beta_\ast)f(\beta_\ast),\phi^{\prime}(\beta_\ast)D\phi^{\prime}(\beta_\ast)^{\top}),\end{equation*}
which is equivalent to 
\begin{equation}\label{split}\sqrt{n}\left(\phi_n(\beta_n)-\beta_\ast\right)+\sqrt{n}\left( \beta_\ast-\phi_n(\beta_\ast) \right) \Rightarrow\mathcal{N}(\phi^{\prime}(\beta_\ast)f(\beta_\ast),\phi^{\prime}(\beta_\ast)D\phi^{\prime}(\beta_\ast)^{\top}).\end{equation}

To make the distribution of $\sqrt{n}\left(\phi_n(\beta_n)-\beta_\ast\right)$, i.e., the first term in the left-hand side of \eqref{split}, converge,  we should require $\sqrt{n}\left( \beta_\ast-\phi_n(\beta_\ast) \right)$, i.e., the second term in the left-hand side of \eqref{split}, has a finite limit. That is to say, the following holds: 
\begin{equation}\label{equivalence1}\phi_n(\beta_\ast)=\beta_\ast+\mathcal{O}\left(\frac{1}{\sqrt{n}}\right).\end{equation}


Since \eqref{equivalence1} holds and $\phi_n$ is differentiable at  $\mathcal{B}(\beta_\ast)$, we can rewrite $\phi_n(\beta_\ast)$ as follows:
\[ \phi_n(\beta_\ast)=\beta_\ast-\frac{1}{\sqrt{n}}g(\beta_\ast)+o\left(\frac{1}{\sqrt{n}}\right),\]
where $g(\beta)$ is differentiable at  $\mathcal{B}(\beta_\ast)$.

In this way, we have that 
\begin{equation}\label{functiongconvergence}\sqrt{n}\left( \phi_n(\beta_\ast)-\beta_\ast \right)=-g(\beta_\ast)+o(1)\end{equation}

In addition, \eqref{equivalence1} indicates $\phi_n^{\prime}(\beta_\ast)\to I$, resulting in the following equivalent reformulation of \eqref{split}:
\begin{equation}\label{split2}\sqrt{n}\left(\phi_n(\beta_n)-\beta_\ast\right)+\sqrt{n}\left( \beta_\ast-\phi_n(\beta_\ast) \right) \Rightarrow\mathcal{N}(f(\beta_\ast),D).\end{equation}

It follows from \eqref{functiongconvergence}, \eqref{split2} and Slutsky’s lemma that
\[\sqrt{n}\left(\phi_n(\beta_n)-\beta_\ast\right)\Rightarrow\mathcal{N}\left(f(\beta_\ast)-g(\beta_\ast),D\right).\]

In this way, the associated  asymptotic mean squared error is \[\tr(D)+\left(f(\beta_\ast)-g(\beta_\ast)\right)^\top\left(f(\beta_\ast)-g(\beta_\ast)\right),\]
implying that the least asymptotic mean squared error is $\tr(D)$ if and only $f(\beta_\ast)=g(\beta_\ast)$.
\end{proof}

\subsection{Proof of Theorem \ref{generaltheorem}}
\begin{proof}
To prove \eqref{generaladjusted}, due to Slutsky's lemma, it suffices to show that  
\[ f_n(\beta_n)-f(\beta_\ast)\to_p 0,\]
which could be guaranteed if 
\begin{equation}\label{eq2}f_n(\beta_\ast)-f(\beta_\ast)\to_p 0,\end{equation}
\begin{equation*}f_n(\beta_n)-f_n(\beta_\ast)\to_p 0,\end{equation*}
where \eqref{eq2} is our assumption. Thus, it suffices to show $f_n(\beta_n)-f_n(\beta_\ast)\to_p 0$ holds.

Since $\sqrt{n}(\beta_n-\beta_\ast)$ converges to some distribution, $\beta_n$ converges to $\beta_\ast$ in probability.
Since $f_n$ is differentiable at $\mathcal{B}(\beta_\ast)$, it follows from the mean value theorem (or Taylor's expansion) that 
\begin{equation*}
\Vert f_n(\beta_n)-f_n(\beta_\ast)\Vert\leq \sup_{ \beta\in\mathcal{B}(\beta_\ast)}\Vert f_n^\prime(\beta)\Vert \Vert \beta_n-\beta_\ast\Vert,
\end{equation*}
It follows from $\beta_n-\beta_\ast\to_p 0$ and $\sup_{\beta\in \mathcal{B}(\beta_\ast)}\Vert f^\prime_n(\beta)\Vert$ is bounded in probability that $f_n(\beta_n)-f_n(\beta_\ast)\to_p0$.
\end{proof}

\subsection{Proof of Theorem \ref{extendeddeltamethod}}
\begin{proof}
The proof of the sequential delta method is based on the proof of the classic delta method, seeing Theorem 3.1 in \cite{van2000asymptotic}.

By the differentiablity of $\phi$ and $\phi_n$, we have the following Taylor's expansions of $\phi_n$ and $\phi$ at $\vartheta$:
\[\phi_n(\theta)-\phi_n(\vartheta)=\phi^\prime_{n}(\vartheta)(\theta-\vartheta)+R_n,\]
\[\phi(\theta)-\phi(\vartheta)=\phi^\prime(\vartheta)(\theta-\vartheta)+R,\]
where $\theta\in\mathcal{B}(\vartheta)$, and $R_n, R$ are associated remainders.
Note that it follows from the stated conditions that $\phi_{n}(\theta)\to \phi(\theta)$, $\phi_{n}(\vartheta)\to \phi(\vartheta)$ and $\phi^\prime_{n}(\vartheta)\to \phi^\prime(\vartheta)$. In this way, we have that $R_n\to R$,
indicating that there exist $N$ such that $\vert R_n\vert\leq 2\vert R\vert$ holds for $\forall n\geq N$.
Since we have that $R=o(\Vert \theta-\vartheta\Vert)$, then $R_n=o(\Vert \theta-\vartheta\Vert)$ holds uniformly for $n\geq N$.

Since the sequence $r_n(T_n-\vartheta)$ converges in distribution, we have that $T_n-\vartheta$ converges to $0$ in probability and $r_n(T_n-\vartheta)$ is uniformly tight.
Then, according to the aforementioned Taylor's expansion, we have that
\begin{equation*}\phi_n(T_n)-\phi_n(\vartheta)
        =\phi^\prime_{n}(\vartheta)(T_n-\vartheta)+ o_p(\Vert T_n-\vartheta\Vert)\end{equation*}
 holds uniformly for $n\geq N$, where $o_p(1)$ means ``converge to $0$ in probability''.  
Then, it follows from the uniform tightness of $r_n(T_n-\vartheta)$ that $o_p(r_n\Vert T_n-\vartheta\Vert)=o_p(1)$. That is to say,
    \begin{equation}\label{taylor} r_n\left(\phi_n(T_n)-\phi_n(\vartheta)\right)
        =r_n\phi^\prime_{n}(\vartheta)(T_n-\vartheta)+ o_p(1),\end{equation}
         holds uniformly for $n\geq N$.

    Because matrix multiplication is continuous and  we have $\phi^\prime_{n}(\vartheta)\to \phi^\prime(\vartheta)$, taking advantage of the extended continuous-mapping theorem, seeing Theorem 1.11.1 in \cite{van1996weak}, we could obtain that
    \begin{equation}\label{extendedcm}r_n\phi_{n}^\prime(\vartheta)(T_n-\vartheta)\Rightarrow \mathcal{N}(\phi^\prime(\vartheta) \mu ,\phi^\prime(\vartheta)\Sigma{\phi^\prime(\vartheta)}^\top).\end{equation}

    Further, it follows from \eqref{taylor}, \eqref{extendedcm} and Slutsky's lemma that
    \[r_n\left(\phi_n(T_n)-\phi_n(\vartheta)\right)\Rightarrow \mathcal{N}(\phi^\prime(\vartheta) \mu ,\phi^\prime(\vartheta)\Sigma{\phi^\prime(\vartheta)}^\top).\]
\end{proof}

\subsection{Proof of Theorem \ref{theorem1}}
\begin{proof}
We denote the inner maximization of the WDRO problem \eqref{formal}, i.e., 
\[\max_{P\in \mathcal{U}_{\rho_n}(\mathbb{P}_n)} \mathbb{E}_{P}[L(f(\mathbf{X},\beta),Y)],\]
by $\Psi_n(\beta)$. 

Then, we have
\begin{equation}\label{vrclass}\Psi_n(\beta)=\inf_{\lambda\geq 0} \left[\lambda\rho^2_n+ \mathbb{E}_{(\mathbf{X},Y)\sim\mathbb{P}_n}\left[\sup_{\mathbf{x}\in\mathbb{R}^d}\left[L(f(\mathbf{x},\beta),Y)-\lambda\Vert \mathbf{x}-\mathbf{X}\Vert_2^2 \right]\right]\right].\end{equation}

Note that Assumption \ref{assume1}, \ref{assumeloss}, and \ref{assumedistr} are extracted from Assumption 1 and 2 in \cite{blanchet2022confidence}, and problem \eqref{vrclass} can be reduced to the problem in Lemma A.1 in \cite{blanchet2022confidence}. Following the same technique, one could derive the convergence in distribution of $\beta_n^{DRO}$:
\[\sqrt{n}(\beta^{DRO}_n-\beta_{\ast})\Rightarrow C(\beta_\ast)^{-1}E-C(\beta_\ast)^{-1}H(\beta_\ast),\]
where \[E\sim \mathcal{N}\left(0,\Cov\left(\frac{\partial L(f(\mathbf{X},\beta),Y)}{\partial \beta}\right)\bigg\vert_{\beta=\beta_{\ast}}\right),\]
\[H(\beta_{\ast})=\tau \frac{\partial \sqrt{\mathbb{E}_{P_\ast}\left[\left\Vert \frac{\partial L(f(\mathbf{X},\beta),Y)}{\partial \mathbf{X}}\right\Vert_2^2\right] }}{\partial \beta}\Bigg\vert_{\beta=\beta_{\ast}}.\]

It follows from the matrix $C(\beta_\ast)$ is positive definite that 
\[\sqrt{n}(\beta^{DRO}_n-\beta_{\ast})\Rightarrow  \mathcal{N}\left(-C(\beta_\ast)^{-1}H(\beta_\ast),C(\beta_\ast)^{-1}\Cov\left(\frac{\partial L(f(\mathbf{X},\beta),Y)}{\partial \beta}\right)\bigg\vert_{\beta=\beta_{\ast}}C(\beta_\ast)^{-1}\right).\]
\end{proof}
\subsection{Proof of Proposition \ref{conditions2}}
\begin{proof}
    Notice we have that 
    \[\frac{\partial^2 L(\langle\mathbf{x},\beta\rangle,y)}{\partial \beta^2}=\frac{\partial^2 L(\langle\mathbf{x},\beta\rangle,y)}{\partial f^2}\mathbf{x}\mathbf{x}^\top. \]
   
    Since $\frac{\partial^2 L(f,y)}{\partial f^2}>0$ and there does not exit nonzero $\alpha$ such that $\mathbb{P}_n(\alpha^{\top}\mathbf{X}=0)=1$, we have that 
    \[\mathbb{E}_{\mathbb{P}_n}\left[\frac{\partial^2 L(f(\mathbf{X},\beta),Y)}{\partial \beta^2}\right]\Bigg\vert_{\beta=\beta_n^{DRO}}\succ 0.\]
    Notice that 
    \begin{equation*}\begin{aligned}
        \left\Vert \frac{\partial L(f(\mathbf{X},\beta),Y)}{\partial \mathbf{X}}\right\Vert_2^2\Bigg\vert_{\beta=\beta_n^{DRO}}=&\left\Vert \frac{\partial L(\langle\mathbf{X},\beta\rangle,Y)}{\partial \mathbf{X}}\right\Vert_2^2\Bigg\vert_{\beta=\beta_n^{DRO}}\\
        =&\left(\frac{\partial L(\langle\mathbf{X},\beta_n^{DRO}\rangle,Y)}{\partial  f}\right)^2\Vert\beta_n^{DRO}\Vert_2^2.    \end{aligned}\end{equation*}
    Since we have $\beta_n^{DRO}\not=\mathbf{0}$ and $\mathbb{P}_n\left( \frac{\partial L(\langle\mathbf{X},\beta_n^{DRO}\rangle,Y)}{\partial  f}\not=0\right)>0$, we have that\
     \[\mathbb{P}_n\left( \left\Vert \frac{\partial L(f(\mathbf{X},\beta),Y)}{\partial \mathbf{X}}\right\Vert_2^2\not=0\right)\Bigg\vert_{ \beta=\beta_n^{DRO}}>0.\]
\end{proof}
\subsection{Proof of Proposition \ref{sufficient2}}
\begin{proof}
Since  $f(\mathbf{x},\beta)=\langle \mathbf{x},\beta\rangle$ holds, we have that
\begin{equation}
\begin{aligned}\label{decomposeH}
H(\beta_{\ast})&=\tau \frac{\partial \sqrt{\mathbb{E}_{P_\ast}\left[\left\Vert \frac{\partial L(\langle \mathbf{X},\beta\rangle,Y)}{\partial \mathbf{X}}\right\Vert_2^2\right] }}{\partial \beta}\Bigg\vert_{\beta=\beta_{\ast}}\\
&=\tau \frac{\partial \left(\Vert \beta\Vert_2 \sqrt{\mathbb{E}_{P_\ast}\left[\left( \frac{\partial L(\langle \mathbf{X},\beta\rangle,Y)}{\partial f}\right)^2\right]}\right)}{\partial \beta}\Bigg\vert_{\beta=\beta_{\ast}}\\
&=\tau\left(\sqrt{\mathbb{E}_{P_\ast}\left[\left( \frac{\partial L(\langle \mathbf{X},\beta_\ast\rangle,Y)}{\partial f}\right)^2\right]}\frac{\beta_\ast}{\Vert\beta_\ast\Vert_2} +\Vert\beta_{\ast}\Vert_2\frac{\mathbb{E}_{P_\ast}\left[ \frac{\partial L(\langle \mathbf{X},\beta_\ast\rangle,Y)}{\partial f}\frac{\partial^2 L(\langle \mathbf{X},\beta_\ast\rangle,Y)}{\partial f^2}\mathbf{X}\right] }{ \sqrt{\mathbb{E}_{P_\ast}\left[\left( \frac{\partial L(\langle \mathbf{X},\beta_\ast\rangle,Y)}{\partial f}\right)^2\right]}}\right).
\end{aligned}\end{equation}

Further, if \[\mathbb{E}_{P_\ast}\left[ \frac{\partial L(\langle \mathbf{X},\beta_\ast\rangle,Y)}{\partial f}\frac{\partial^2 L(\langle \mathbf{X},\beta_\ast\rangle,Y)}{\partial f^2}\mathbf{X}\right]=0\] holds, 
the second term in the equation \eqref{decomposeH} equals to 0.

Then, we have
\[H(\beta_\ast)=\tau\sqrt{\mathbb{E}_{P_\ast}\left[\left( \frac{\partial L(\langle \mathbf{X},\beta\rangle,Y)}{\partial f}\right)^2\right]}\Bigg\vert_{\beta=\beta_\ast}\frac{\beta_\ast}{{\Vert\beta_\ast\Vert_2}}.\] 
\end{proof}

\subsection{Proof of Theorem \ref{theorem2}}
\begin{proof}
Note we have that 
\[\frac{\partial \sqrt{\mathbb{E}\left[\left\Vert \frac{\partial L(f(\mathbf{X},\beta),Y)}{\partial \mathbf{X}}\right\Vert_2^2\right] }}{\partial \beta}
=\frac{\mathbb{E}\left[ \frac{\partial L(f(\mathbf{X},\beta),Y)}{\partial \mathbf{X}\partial\beta}\frac{\partial L(f(\mathbf{X},\beta),Y)}{\partial \mathbf{X}}\right]}{\sqrt{\mathbb{E}\left[\left\Vert \frac{\partial L(f(\mathbf{X},\beta),Y)}{\partial \mathbf{X}}\right\Vert_2^2\right] }}.\]

In this way, we have that \[f(\mathbf{z})=-C(\mathbf{z})^{-1}H(\mathbf{z})\] 
    \[ f_n(\mathbf{z})=-C_n(\mathbf{z})^{-1} H_n(\mathbf{z}), \]
    where 
    \[C(\mathbf{z})=\mathbb{E}_{P_{\ast}}\left[\frac{\partial^2 L(f(\mathbf{X},\beta),Y)}{\partial \beta^2}\right]\Bigg\vert_{\beta=\mathbf{z}}, \quad H(\mathbf{z})=\tau\frac{\mathbb{E}_{P_\ast}\left[ \frac{\partial L(f(\mathbf{X},\beta),Y)}{\partial \mathbf{X}\partial\beta}\frac{\partial L(f(\mathbf{X},\beta),Y)}{\partial \mathbf{X}}\right]}{\sqrt{\mathbb{E}_{P_\ast}\left[\left\Vert \frac{\partial L(f(\mathbf{X},\beta),Y)}{\partial \mathbf{X}}\right\Vert_2^2\right] }}\Bigg\vert_{\beta=\mathbf{z}},\]
    \[ C_n(\mathbf{z})=\mathbb{E}_{\mathbb{P}_n}\left[\frac{\partial^2 L(f(\mathbf{X},\beta),Y)}{\partial \beta^2}\right]\Bigg\vert_{\beta=\mathbf{z}}, \quad H_n(\mathbf{z})=\tau\frac{\mathbb{E}_{\mathbb{P}_n}\left[ \frac{\partial L(f(\mathbf{X},\beta),Y)}{\partial \mathbf{X}\partial\beta}\frac{\partial L(f(\mathbf{X},\beta),Y)}{\partial \mathbf{X}}\right]}{\sqrt{\mathbb{E}_{\mathbb{P}_n}\left[\left\Vert \frac{\partial L(f(\mathbf{X},\beta),Y)}{\partial \mathbf{X}}\right\Vert_2^2\right] }}\Bigg\vert_{\beta=\mathbf{z}}.\]
    
It follows from Theorem \ref{generaltheorem} that it suffices to show $f_n$ satisfies Assumption \ref{assumefn}. 

It follows from Assumption \ref{assumeloss} that $L(f(\mathbf{x},\beta),y)$ is twice differentiable, $\frac{\partial L(f(\mathbf{x},\beta),y)}{\partial \mathbf{x}\partial \beta}$ and $\frac{\partial^2 L(f(\mathbf{x},\beta),y)}{\partial \beta^2}$ are differentiable w.r.t. $\beta$, indicating that both $f_n(\mathbf{z})=-C_n(\mathbf{z})^{-1}H_n(\mathbf{z})$ and $f(\mathbf{z})=-C(\mathbf{z})^{-1}H(\mathbf{z})$ are differentiable at $\mathcal{B}(\beta_\ast)$. The first item in Assumption \ref{assumefn} is satisfied.

Notably, since $L(f(\mathbf{x},\beta),y)$ is twice continuously differentiable, and $\frac{\partial L(f(\mathbf{x},\beta),y)}{\partial \mathbf{x}\partial \beta}$, $\frac{\partial^2 L(f(\mathbf{x},\beta),y)}{\partial \beta^2}$  are continuously differentiable w.r.t $\beta$, then the the gradient of  $f(\mathbf{z})=-C(\mathbf{z})^{-1}H(\mathbf{z})$, i.e., $f^\prime(\mathbf{z})$, is continuous at  $\mathcal{B}(\beta_\ast)$. In this way, we have that $\sup_{\beta\in\mathcal{B}(\beta_\ast)}\Vert f^\prime(\beta)\Vert$ is bounded. In addition, the law of large numbers implies $f_n^\prime(\mathbf{z})\to_pf^\prime(\mathbf{z})$ holds for every $\mathbf{z}$ at  $\mathcal{B}(\beta_\ast)$. This convergence promises that $\sup_{\beta\in\mathcal{B}(\beta_\ast)}\Vert f_n^\prime(\beta)\Vert$ is bounded in probability. The second item in Assumption \ref{assumefn} is satisfied.

Since $C_n(\mathbf{z})$ and $H_n(\mathbf{z})$ are defined in terms of the empirical distribution, $f_n(\beta_\ast)\to_pf(\beta_\ast)$ holds due to the law of large numbers. The third item in Assumption \ref{assumefn} is satisfied.

\end{proof}

\subsection{Proof of Corollary \ref{coro1}}
Since the loss function $L(f(\mathbf{x},\beta),y)$ is $h$-Lipschitz continuous w.r.t. $\beta$, we have that 
\[\vert L(f(\mathbf{x},\beta_n^{DRO}),y)-L(f(\mathbf{x},\beta_n^{ADRO}),y)\vert \leq  h\Vert\beta_n^{DRO} -\beta_n^{ADRO}\Vert_2,\]
indicating
\[\mathbb{E}_{P_\ast}\left[L(f(\mathbf{X},\beta_n^{ADRO}),Y)\right]- h\mathbb{E}_{P_\ast}\left[\Vert\beta_n^{DRO} -\beta_n^{ADRO}\Vert_2\right] \leq \mathbb{E}_{P_\ast}\left[L(f(\mathbf{X},\beta_n^{DRO}),Y)\right],\]
and 
\begin{equation*}\begin{aligned}
    &\sup_{P\in\mathcal{U}_{\rho_n}(\mathbb{P}_n)}\mathbb{E}_{P}\left[L(f(\mathbf{X},\beta_n^{DRO}),Y)\right]\\
    &\leq 
 \sup_{P\in\mathcal{U}_{\rho_n}(\mathbb{P}_n)}\mathbb{E}_{P}\left[L(f(\mathbf{X},\beta_n^{ADRO}),Y)\right]+h\sup_{P\in\mathcal{U}_{\rho_n}(\mathbb{P}_n)}\mathbb{E}_{P}\left[\Vert\beta_n^{DRO} -\beta_n^{ADRO}\Vert_2\right].\end{aligned} \end{equation*}

Since we have the following definition of $\beta_n^{ADRO}$:
\[\beta_n^{ADRO}=\beta_n^{DRO}+\frac{C_n(\beta_n^{DRO})^{-1}H_n(\beta_n^{DRO})}{\sqrt{n}},\]
together with \eqref{outofsample}, we have that 
\begin{equation*}
\begin{aligned}
&\mathbb{E}_{P_{\ast}}\left[ L(f(\mathbf{X},\beta_n^{ADRO}),Y)\right]\leq\sup_{P\in \mathcal{U}_{\rho_n}(\mathbb{P}_n)} \mathbb{E}_{P} \left[L(f(\mathbf{X},\beta_n^{ADRO}),Y)\right]\\
&+ \frac{h}{\sqrt{n}}\left( 
\mathbb{E}_{P_\ast}\left[\Vert C_n(\beta_n^{DRO})^{-1}H_n(\beta_n^{DRO})\Vert_2\right]+ \sup_{P\in\mathcal{U}_{\rho_n}(\mathbb{P}_n)}\mathbb{E}_P\left[\Vert C_n(\beta_n^{DRO})^{-1}H_n(\beta_n^{DRO})\Vert_2\right]
\right)+\epsilon_n, 
\end{aligned}
\end{equation*}
holds with probability $1-\alpha$.
\subsection{Proof of Lemma \ref{checklemma1}}
\begin{proof}
    \textbf{a.} The loss function $L^{1}\left(\langle \mathbf{x},\beta\rangle,y\right)=\log(1+e^{-y\langle \mathbf{x},\beta\rangle})$ is twice continuously differentiable w.r.t. $\mathbf{x}$ and $\beta$.

    \textbf{b.} Since we have that \[\frac{\partial^2 L^{1}\left(\langle \mathbf{x},\beta\rangle,y\right)}{\partial \beta^2}=\frac{e^{y\langle\mathbf{x},\beta\rangle}\mathbf{x}\mathbf{x}^{\top}}{\left(1+e^{y\langle\mathbf{x},\beta\rangle}\right)^2}\succeq 0,\]
    where $\succeq$ means the matrix is positive semidefinite,
    the function $L^{1}\left(\langle \mathbf{x},\beta\rangle,y\right)$ is convex w.r.t. $\beta$.
    
     \textbf{c.} Note we have that
    \begin{equation*}
    \begin{aligned}
    \left\Vert\frac{\partial^2 L^{1}\left(\langle \mathbf{x},\beta\rangle,y\right)}{\partial \mathbf{x}^2}\right\Vert_2&=\left\Vert\frac{\beta\beta^{\top}e^{y\langle\mathbf{x},\beta\rangle}}{\left(1+e^{y\langle\mathbf{x},\beta\rangle}\right)^2}\right\Vert_2\\
    &=\left\Vert \beta\right\Vert_2^2 \frac{e^{y\langle\mathbf{x},\beta\rangle}}{\left(1+e^{y\langle\mathbf{x},\beta\rangle}\right)^2}<M(\beta)=\Vert\beta\Vert_2^2.
    \end{aligned}
    \end{equation*}
   
    Further, we have that
    \[\frac{\partial \left\Vert\frac{\partial^2 L^{1}\left(\langle \mathbf{x},\beta\rangle,y\right)}{\partial \mathbf{x}^2}\right\Vert_2}{\partial \mathbf{x}}=\left\Vert \beta\right\Vert_2^2 \frac{y e^{y\langle \mathbf{x},\beta\rangle} \left( 1-e^{y\langle \mathbf{x},\beta\rangle}\right)}{\left(1+e^{y\langle\mathbf{x},\beta\rangle}\right)^3}\beta.\]
  We know that
    $\frac{ e^{y\langle \mathbf{x},\beta\rangle} \left( 1-e^{y\langle \mathbf{x},\beta\rangle}\right)}{\left(1+e^{y\langle\mathbf{x},\beta\rangle}\right)^3}$
    is bounded. Since $\beta\in B$ and $B$ is bounded, we have that $\frac{\partial \left\Vert\frac{\partial^2L^{1}\left(\langle \mathbf{x},\beta\rangle,y\right)}{\partial \mathbf{x}^2}\right\Vert_2}{\partial \mathbf{x}}$ is bounded, implying $\left\Vert\frac{\partial^2 L^{1}\left(\langle \mathbf{x},\beta\rangle,y\right)}{\partial \mathbf{x}^2}\right\Vert_2$ is uniformly continuous w.r.t. $\mathbf{x}$.
\end{proof}

\subsection{Proof of Lemma \ref{poissonassume}}
\begin{proof}
    \textbf{a.} The loss function $L^{2}\left(\langle\mathbf{x},\beta\rangle,y\right)=e^{\langle \mathbf{x},\beta\rangle}-y\langle \mathbf{x},\beta\rangle$ is twice continuously differentiable w.r.t. $\mathbf{x}$ and $\beta$.

\textbf{b.} Because we have\[\frac{\partial^2 L^{2}(\langle\mathbf{x},\beta\rangle,y)}{\partial \beta^2}=e^{\langle \mathbf{x},\beta\rangle}\mathbf{x}\mathbf{x}^{\top}\succeq 0,\]
the function $L^{2}\left(\langle\mathbf{x},\beta\rangle,y\right)$ is convex w.r.t. $\beta$.

\textbf{c.}
We have
\[\left\Vert\frac{\partial^2 L^{2}(\langle\mathbf{x},\beta\rangle,y)}{\partial \mathbf{x}^2}\right\Vert_2 = \left\Vert\beta\beta^{\top}\right\Vert_2 e^{\langle \mathbf{x},\beta\rangle}=\Vert \beta\Vert_2^2 e^{\langle\mathbf{x},\beta\rangle}.\]

Since $\mathbf{x}\in \Omega, \beta\in B$, where both $\Omega$ and $B$ are bounded, $\Vert\frac{\partial^2 L^{2}(\langle\mathbf{x},\beta\rangle,y)}{\partial \mathbf{x}^2}\Vert_2$ is bounded by a function of $\beta$ and uniformly continuous w.r.t. $\mathbf{x}$.
\end{proof}

\subsection{Proof of Lemma \ref{linearassume}}
\begin{proof}
    \textbf{a.} The loss function $L^{3}(\langle\mathbf{x},\beta\rangle,y)=\frac{1}{2}\left(\langle\mathbf{x},\beta\rangle-y\right)^2$ is twice continuously differentiable w.r.t. $\mathbf{x}$ and $\beta$.

\textbf{b.} The loss function $L^{3}(\langle\mathbf{x},\beta\rangle,y)=\frac{1}{2}\left(\langle\mathbf{x},\beta\rangle-y\right)^2$ is convex w.r.t. $\beta$.

     \textbf{c.} We have
      \[\left\Vert\frac{\partial^2 L(f(\mathbf{x},\beta),y)}{\partial \mathbf{x}^2}\right\Vert_2=\Vert 2\beta\beta^{\top}\Vert_2=2\Vert\beta\Vert_2^2.\]
      
      Since $\beta\in B$ and $B$ is bounded, $\Vert\frac{\partial^2 L(f(\mathbf{x},\beta),y)}{\partial \mathbf{x}^2}\Vert_2$ is bounded by  function of $2\Vert\beta\Vert_2^2$ and uniformly continuous w.r.t. $\mathbf{x}$.
\end{proof}
\subsection{Proof of Lemma \ref{checklemma2}}
\begin{proof}
\textbf{a.}
From the equation
\[ \frac{\partial L^{1}\left(\langle \mathbf{x},\beta\rangle,y\right)}{\partial \beta}=\frac{-y\mathbf{x}}{1+e^{y\langle \mathbf{x},\beta\rangle}},\]
and the assumption $\mathbb{E}_{P_\ast}\left[\Vert \mathbf{X}\Vert_2^2\right]<\infty$, we have that
\begin{equation*}\begin{aligned}\mathbb{E}_{P_{\ast}} \left[\left\Vert \frac{\partial L^{1}(\langle\mathbf{X},\beta\rangle,Y)}{\partial \beta}\right\Vert_2^2\right]\Bigg\vert_{\beta=\beta_{\ast}}&=\mathbb{E}_{P_\ast}\left[\frac{\Vert \mathbf{X}\Vert_2^2}{(1+e^{Y\langle \mathbf{X},\beta_\ast\rangle})^2}\right]\\
&<\mathbb{E}_{P_\ast}\left[\Vert \mathbf{X}\Vert_2^2\right]<\infty.\end{aligned}\end{equation*}

Since we have that
\[\frac{\partial^2 L^{1}(\langle\mathbf{x},\beta\rangle,y)}{\partial \beta^2}=\frac{e^{y\langle\mathbf{x},\beta\rangle}\mathbf{x}\mathbf{x}^{\top} }{\left( 1+e^{y\langle \mathbf{x},\beta\rangle}\right)^2},\]
 where \[e^{y\langle\mathbf{x},\beta\rangle}/ (1+e^{y\langle \mathbf{x},\beta\rangle})^2>0,\] and there does not exist nonzero $\alpha$ such that $P_{\ast}(\alpha^{\top}\mathbf{X}=0)=1$, then we could conclude
\[\mathbb{E}_{P_{\ast}}\left[\frac{\partial^2 L^{1}(\langle\mathbf{X},\beta\rangle,Y)}{\partial \beta^2}\right]\Bigg\vert_{\beta=\beta_{\ast}}\succ 0.\]

In addition, we have that
    \begin{equation*}\begin{aligned}
    &\mathbb{E}_{P_{\ast}}\left[\frac{\partial L^{1}(\langle\mathbf{X},\beta\rangle,Y)}{\partial\beta}\right]\Bigg\vert_{\beta=\beta_{\ast}}\\
    =&\mathbb{E}_{P_{\ast}}\left[ \frac{-Y\mathbf{X}}{1+e^{Y\langle\mathbf{X},\beta_\ast\rangle}}\right]\\
    =&\int P_\ast(Y=1|\mathbf{X}=\mathbf{x})\frac{\mathbf{x}}{1+e^{\langle\mathbf{x},\beta_\ast\rangle}}dF_\ast(\mathbf{x})+\int P_\ast(Y=-1|\mathbf{X}=\mathbf{x})\frac{\mathbf{x}}{1+e^{-\langle\mathbf{x},\beta_\ast\rangle}}dF_\ast(\mathbf{x})\\
    =&\int \frac{\mathbf{x}}{1+e^{-\langle\mathbf{x},\beta_\ast\rangle}}\frac{-1}{1+e^{\langle\mathbf{x},\beta_\ast\rangle}}dF_\ast(\mathbf{x})+\int \frac{\mathbf{x}}{1+e^{\langle\mathbf{x},\beta_\ast\rangle}}\frac{1}{1+e^{-\langle\mathbf{x},\beta_\ast\rangle}}dF_\ast(\mathbf{x})\\
    =&0,
    \end{aligned}\end{equation*}
 where $F_\ast$ is the distribution function of $P_\ast$.
 
\textbf{b.} Notice we have that
\[\frac{\partial L^{1}(\langle\mathbf{x},\beta\rangle,y)}{\partial \mathbf{x}}\Bigg\vert_{\beta=\beta_{\ast}}=\frac{-y\beta_{\ast}}{1+e^{y\langle\mathbf{x},\beta_\ast\rangle}},\]
where \[\beta_{\ast}\not=\mathbf{0}, y\not=0,1+e^{y\langle\mathbf{x},\beta_\ast\rangle}>0, \] then we can conclude that
\[P_{\ast}\left(\frac{\partial L^{1}(\langle\mathbf{X},\beta\rangle,Y)}{\partial \mathbf{X}}\not =0\right)\bigg\vert_{\beta=\beta_{\ast}}>0.\]

Then, we have that
    \[\frac{\partial ^2L^{1}(\langle\mathbf{x},\beta\rangle,Y)}{\partial \mathbf{x}\partial \beta}\Bigg\vert_{\beta=\beta_{\ast}}=\frac{-yI_d}{1+e^{y\langle \mathbf{x},\beta_\ast\rangle}}+\frac{e^{y\langle\mathbf{x},\beta_\ast\rangle}\beta_\ast\mathbf{x}^{\top} }{\left( 1+e^{y\langle \mathbf{x},\beta_\ast\rangle}\right)^2}.\]

Since the kernel space of the matrix $\frac{\partial ^2L^{1}(\langle\mathbf{x},\beta\rangle,Y)}{\partial \mathbf{x}\partial \beta}\big\vert_{\beta=\beta_{\ast}}$ is different for different $\mathbf{x},y$, we can conclude that 
\[ \mathbb{E}_{P_{\ast}}\left[\frac{\partial ^2L^{1}(\langle\mathbf{X},\beta\rangle,Y)}{\partial \mathbf{X}\partial \beta}\left(\frac{\partial ^2L^{1}(\langle\mathbf{X},\beta\rangle,Y)}{\partial \mathbf{X}\partial \beta}\right)^{\top}\right]\Bigg\vert_{\beta=\beta_{\ast}}\succ 0.\]
\end{proof}
\subsection{Proof of Lemma \ref{poissonassume2}}
\begin{proof}
\textbf{a.}
From the equation
\[ \frac{\partial L^{2}\left(\langle \mathbf{x},\beta\rangle,y\right)}{\partial \beta}=\mathbf{x}e^{\langle \mathbf{x},\beta\rangle}-y\mathbf{x},\]
 we have that
\begin{equation*}
\begin{aligned}
&\mathbb{E}_{P_{\ast}} \left[\left\Vert \frac{\partial L^{2}(\langle\mathbf{X},\beta\rangle,Y)}{\partial \beta}\right\Vert_2^2\right]\Bigg\vert_{\beta=\beta_{\ast}}\\
=&\mathbb{E}_{P_\ast}\left[ \Vert\mathbf{X}\Vert_2^2 \left( e^{\langle \mathbf{X},\beta_\ast\rangle}-Y\right)^2 \right]\\
=&\mathbb{E}_{P_{\ast}}\left[\Vert\mathbf{X}\Vert_2^2\mathbb{E}_{P_\ast}\left[ \left( e^{\langle \mathbf{X},\beta_\ast\rangle}-Y\right)^2\Big\vert \mathbf{X}\right]\right]\\
\end{aligned}
\end{equation*}   

Since $Y\vert\mathbf{X}=\mathbf{x}$ follows the Poisson distribution with parameter $e^{\langle\mathbf{x},\beta_\ast\rangle}$, we have that 
\begin{equation*}\begin{aligned}
    \mathbb{E}_{P_{\ast}} \left[\left\Vert \frac{\partial L^{2}(\langle\mathbf{X},\beta\rangle,Y)}{\partial \beta}\right\Vert_2^2\right]\Bigg\vert_{\beta=\beta_{\ast}}&=\mathbb{E}_{P_{\ast}}\left[\Vert\mathbf{X}\Vert_2^2\Var_{P_\ast}(Y\vert\mathbf{X})\right]\\
    &=\mathbb{E}_{P_{\ast}}\left[ \Vert\mathbf{X}\Vert_2^2e^{\langle\mathbf{X},\beta_\ast\rangle} \right]<\infty.
    \end{aligned}\end{equation*}

Since we have that
\[\frac{\partial^2 L^{2}(\langle\mathbf{x},\beta\rangle,y)}{\partial \beta^2}=e^{\langle\mathbf{x},\beta\rangle}\mathbf{x}\mathbf{x}^{\top} ,\]
 where $e^{\langle\mathbf{x},\beta\rangle}>0$, and there does not exist nonzero $\alpha$ such that $P_{\ast}(\alpha^{\top}\mathbf{X}=0)=1$, we could conclude that
\[\mathbb{E}_{P_{\ast}}\left[\frac{\partial^2 L^{2}(\langle\mathbf{X},\beta\rangle,Y)}{\partial \beta^2}\right]\Bigg\vert_{\beta=\beta_{\ast}}\succ 0.\]

In addition, we have that
  \begin{equation*}\begin{aligned}
    &\mathbb{E}_{P_{\ast}}\left[\frac{\partial L^{2}(\langle\mathbf{X},\beta\rangle,Y)}{\partial\beta}\right]\Bigg\vert_{\beta=\beta_{\ast}}\\=&\mathbb{E}_{P_{\ast}}\left[ \mathbf{X} e^{\langle\mathbf{X},\beta_{\ast}\rangle}-Y\mathbf{X}\right]\\
    =&\mathbb{E}_{P_{\ast}}\left[e^{\langle\mathbf{X},\beta_{\ast}\rangle} \mathbf{X} -\mathbb{E}_{P_{\ast}}\left[Y\vert \mathbf{X}\right]\mathbf{X}\right]\\
    =&0.
    \end{aligned}\end{equation*}
    \textbf{b.} Notice we have that
\[\frac{\partial L^{2}(\langle\mathbf{x},\beta\rangle,y)}{\partial \mathbf{x}}\Bigg\vert_{\beta=\beta_{\ast}}=(e^{\langle \mathbf{x},\beta_\ast\rangle}-y)\beta_\ast,\]
where $\beta_{\ast}\not=\mathbf{0}$,
\[P_\ast\left(e^{\langle \mathbf{X},\beta_\ast\rangle}-Y\not=0\right)>0,\]
then we can conclude that
\[P_{\ast}\left(\frac{\partial L^{2}(\langle\mathbf{X},\beta\rangle,Y)}{\partial \mathbf{X}}\not =0\right)\bigg\vert_{\beta=\beta_{\ast}}>0.\]

Then, we have that
    \[\frac{\partial ^2L^{2}(\langle\mathbf{x},\beta\rangle,Y)}{\partial \mathbf{x}\partial \beta}\Bigg\vert_{\beta=\beta_{\ast}}=( e^{\langle\mathbf{x},\beta_\ast\rangle}-y)I_d+e^{\langle\mathbf{x},\beta_\ast\rangle}\beta_\ast \mathbf{x}^{\top} .\]

Since the kernel space of the matrix $\frac{\partial ^2L^{2}(\langle\mathbf{x},\beta\rangle,Y)}{\partial \mathbf{x}\partial \beta}\big\vert_{\beta=\beta_{\ast}}$ is different for different $\mathbf{x},y$, then we can conclude that 
\[ \mathbb{E}_{P_{\ast}}\left[\frac{\partial ^2L^{2}(\langle\mathbf{X},\beta\rangle,Y)}{\partial \mathbf{X}\partial \beta}\left(\frac{\partial ^2L^{2}(\langle\mathbf{X},\beta\rangle,Y)}{\partial \mathbf{X}\partial \beta}\right)^{\top}\right]\Bigg\vert_{\beta=\beta_{\ast}}\succ 0.\]
\end{proof}
\subsection{Proof of Lemma \ref{linearassume2}}
\begin{proof}
\textbf{a.}    From the equation
\[ \frac{\partial L^{3}\left(\langle \mathbf{x},\beta\rangle,y\right)}{\partial \beta}=(\langle\mathbf{x},\beta\rangle-y)\mathbf{x},\]
 we have that
\begin{equation*}
\begin{aligned}
&\mathbb{E}_{P_{\ast}} \left[\left\Vert \frac{\partial L^{3}(\langle\mathbf{X},\beta\rangle,Y)}{\partial \beta}\right\Vert_2^2\right]\Bigg\vert_{\beta=\beta_{\ast}}\\
=&\mathbb{E}_{P_\ast}\left[ \Vert\mathbf{X}\Vert_2^2 \left( \langle \mathbf{X},\beta_\ast\rangle-Y\right)^2 \right]\\
=&\mathbb{E}_{P_{\ast}}\left[\Vert\mathbf{X}\Vert_2^2 \mathbb{E}_{P_\ast}\left[ \left( \langle \mathbf{X},\beta_\ast\rangle-Y\right)^2\Big\vert \mathbf{X}\right]\right]\\
\end{aligned}
\end{equation*} 

Notice that $Y\vert \mathbf{X}=\mathbf{x}$ follows the normal distribution with a mean value of $\langle \mathbf{x},\beta_\ast\rangle$.
Thus, we have that
\begin{equation}\begin{aligned}
    \mathbb{E}_{P_{\ast}} \left[\left\Vert \frac{\partial L^{3}(\langle\mathbf{X},\beta\rangle,Y)}{\partial \beta}\right\Vert_2^2\right]\Bigg\vert_{\beta=\beta_{\ast}}&=\mathbb{E}_{P_\ast}\left[ \Vert\mathbf{X}\Vert_2^2\Var_{P_\ast}(Y|\mathbf{X})\right]<\infty.
\end{aligned}\end{equation}

Since we have that
\[\frac{\partial^2 L^{3}(\langle\mathbf{x},\beta\rangle,y)}{\partial \beta^2}=\mathbf{x}\mathbf{x}^{\top},\]
and there does not exist nonzero $\alpha$ such that $P_{\ast}(\alpha^{\top}\mathbf{X}=0)=1$, we could conclude that
\[\mathbb{E}_{P_{\ast}}\left[\frac{\partial^2 L^{3}(\langle\mathbf{X},\beta\rangle,Y)}{\partial \beta^2}\right]\Bigg\vert_{\beta=\beta_\ast}\succ 0.\]

In addition, we have that
  \begin{equation*}\begin{aligned}
    &\mathbb{E}_{P_{\ast}}\left[\frac{\partial L^{3}(\langle\mathbf{X},\beta\rangle,Y)}{\partial\beta}\right]\Bigg\vert_{\beta=\beta_{\ast}}\\
    =&\mathbb{E}_{P_{\ast}}\left[\langle\mathbf{X},\beta_{\ast}\rangle \mathbf{X} -Y\mathbf{X}\right]\\
    =&\mathbb{E}_{P_{\ast}}\left[ \langle\mathbf{X},\beta_{\ast}\rangle\mathbf{X} -\mathbb{E}_{P_{\ast}}\left[Y\vert\mathbf{X}\right]\mathbf{X}\right]\\
    =&0.
    \end{aligned}\end{equation*}
    \textbf{b.}
     Notice that,
\[\frac{\partial L^{3}(\langle\mathbf{x},\beta\rangle,y)}{\partial \mathbf{x}}\Bigg\vert_{\beta=\beta_{\ast}}= \left(\langle \mathbf{x},\beta_\ast\rangle-y\right)\beta_\ast,\]
where $\beta_{\ast}\not=\mathbf{0}$,
\[P_\ast\left(\langle \mathbf{X},\beta_\ast\rangle-Y\not=0\right)>0,\]
then we can conclude that
\[P_{\ast}\left(\frac{\partial L^{3}(\langle\mathbf{X},\beta\rangle,Y)}{\partial \mathbf{X}}\not =0\right)\bigg\vert_{\beta=\beta_{\ast}}>0.\]

Then, we have that
    \[\frac{\partial ^2L^{3}(\langle\mathbf{x},\beta\rangle,Y)}{\partial \mathbf{x}\partial \beta}\Bigg\vert_{\beta=\beta_{\ast}}=\left( \langle\mathbf{x},\beta_\ast\rangle-y\right)I_d+\beta_\ast \mathbf{x}^{\top}.\]

 Since the kernel space of the matrix $\frac{\partial ^2L^{3}(\langle\mathbf{x},\beta\rangle,Y)}{\partial \mathbf{x}\partial \beta}\big\vert_{\beta=\beta_{\ast}}$ is different for different $\mathbf{x}, y$, we can conclude that 
\[ \mathbb{E}_{P_{\ast}}\left[\frac{\partial ^2L^{3}(\langle\mathbf{X},\beta\rangle,Y)}{\partial \mathbf{X}\partial \beta}\left(\frac{\partial ^2L^{3}(\langle\mathbf{X},\beta\rangle,Y)}{\partial \mathbf{X}\partial \beta}\right)^{\top}\right]\Bigg\vert_{\beta=\beta_{\ast}}\succ 0.\]
\end{proof}
\subsection{Proof of Proposition \ref{logistic}}\label{prooflogistic}
\begin{proof}
Regarding the asymptotic covariance matrix, since we have that
    \begin{equation*}
    \begin{aligned}&\Cov_{P_\ast}\left(\frac{\partial L^{1}(\langle\mathbf{X},\beta\rangle,Y)}{\partial \beta}\right)\bigg\vert_{\beta=\beta_{\ast}}\\
    &=\mathbb{E}_{P_\ast}\left[ \frac{\partial L^{1}(\langle\mathbf{X},\beta\rangle,Y)}{\partial \beta}\left(\frac{\partial L^{1}(\langle\mathbf{X},\beta\rangle,Y)}{\partial \beta}\right)^\top\right]\\&=\mathbb{E}_{P_{\ast}}\left[ \frac{\mathbf{X}\mathbf{X}^{\top}}{\left( 1+e^{Y\langle\mathbf{X},\beta_{\ast}\rangle}\right)^2}\right]\\
    &=\int P_\ast(Y=1|\mathbf{X}=\mathbf{x}) \frac{\mathbf{x}\mathbf{x}^\top}{\left(1+e^{\langle \mathbf{x},\beta_{\ast}\rangle}\right)^2}dF_\ast(\mathbf{x}) +\int P_\ast(Y=-1|\mathbf{X}=\mathbf{x}) \frac{\mathbf{x}\mathbf{x}^\top}{\left(1+e^{-\langle \mathbf{x},\beta_{\ast}\rangle}\right)^2}dF_\ast(\mathbf{x})\\
    &=\int \frac{1}{1+e^{-\langle \mathbf{x},\beta_{\ast}\rangle}} \frac{\mathbf{x}\mathbf{x}^\top}{\left(1+e^{\langle \mathbf{x},\beta_{\ast}\rangle}\right)^2}dF_\ast(\mathbf{x}) +\int \frac{1}{1+e^{\langle \mathbf{x},\beta_{\ast}\rangle}}\frac{\mathbf{x}\mathbf{x}^\top}{\left(1+e^{-\langle \mathbf{x},\beta_{\ast}\rangle}\right)^2}dF_\ast(\mathbf{x})\\
    &=\int  \frac{e^{\langle \mathbf{x},\beta_\ast\rangle}\mathbf{x}\mathbf{x}^\top}{\left(1+e^{\langle \mathbf{x},\beta_{\ast}\rangle}\right)^3}dF_\ast(\mathbf{x}) +\int\frac{e^{2\langle \mathbf{x},\beta_\ast\rangle}\mathbf{x}\mathbf{x}^\top}{\left(1+e^{\langle \mathbf{x},\beta_{\ast}\rangle}\right)^3}dF_\ast(\mathbf{x})\\
    &=\int\frac{e^{\langle \mathbf{x},\beta_\ast\rangle}\mathbf{x}\mathbf{x}^\top}{\left(1+e^{\langle \mathbf{x},\beta_{\ast}\rangle}\right)^2}dF_\ast(\mathbf{x})\\
    &=\mathbb{E}_{P_{\ast}}\left[ \frac{e^{\langle\mathbf{X},\beta_\ast\rangle}\mathbf{X}\mathbf{X}^{\top}}{\left( 1+e^{\langle\mathbf{X},\beta_{\ast}\rangle}\right)^2}\right],
    \end{aligned}\end{equation*}
    and    \begin{equation*}
    \begin{aligned}&C(\beta_\ast)=\mathbb{E}_{P_{\ast}}\left[\frac{\partial^2 L^{1}(\langle\mathbf{X},\beta\rangle,Y)}{\partial \beta^2}\right]\\
    &=\mathbb{E}_{P_{\ast}}\left[ \frac{\mathbf{X}\mathbf{X}^{\top}e^{Y\langle\mathbf{X},\beta_{\ast}\rangle}}{\left( 1+e^{Y\langle\mathbf{X},\beta_{\ast}\rangle}\right)^2}\right]\\
    &=\int P(Y=1|\mathbf{X}=\mathbf{x}) \frac{e^{\langle \mathbf{x},\beta_{\ast}\rangle}\mathbf{x}\mathbf{x}^\top }{\left(1+e^{\langle \mathbf{x},\beta_{\ast}\rangle}\right)^2}dF_\ast(\mathbf{x}) +\int P(Y=-1|\mathbf{X}=\mathbf{x}) \frac{e^{-\langle \mathbf{x},\beta_{\ast}\rangle}\mathbf{x}\mathbf{x}^\top }{\left(1+e^{-\langle \mathbf{x},\beta_{\ast}\rangle}\right)^2}dF_\ast(\mathbf{x})\\
    &=\int \frac{1}{1+e^{-\langle \mathbf{x},\beta_{\ast}\rangle}} \frac{e^{\langle \mathbf{x},\beta_{\ast}\rangle}\mathbf{x}\mathbf{x}^\top }{\left(1+e^{\langle \mathbf{x},\beta_{\ast}\rangle}\right)^2}dF_\ast(\mathbf{x}) +\int \frac{1}{1+e^{\langle \mathbf{x},\beta_{\ast}\rangle}}\frac{e^{-\langle \mathbf{x},\beta_{\ast}\rangle}\mathbf{x}\mathbf{x}^\top }{\left(1+e^{-\langle \mathbf{x},\beta_{\ast}\rangle}\right)^2}dF_\ast(\mathbf{x})\\
    &=\int  \frac{e^{2\langle \mathbf{x},\beta_\ast\rangle}\mathbf{x}\mathbf{x}^\top}{\left(1+e^{\langle \mathbf{x},\beta_{\ast}\rangle}\right)^3}dF_\ast(\mathbf{x}) +\int\frac{e^{\langle \mathbf{x},\beta_\ast\rangle}\mathbf{x}\mathbf{x}^\top}{\left(1+e^{\langle \mathbf{x},\beta_{\ast}\rangle}\right)^3}dF_\ast(\mathbf{x})\\
    &=\int\frac{e^{\langle \mathbf{x},\beta_\ast\rangle}\mathbf{x}\mathbf{x}^\top}{\left(1+e^{\langle \mathbf{x},\beta_{\ast}\rangle}\right)^2}dF_\ast(\mathbf{x})\\
    &=\mathbb{E}_{P_{\ast}}\left[ \frac{e^{\langle\mathbf{X},\beta_\ast\rangle}\mathbf{X}\mathbf{X}^{\top}}{\left( 1+e^{\langle\mathbf{X},\beta_{\ast}\rangle}\right)^2}\right],
    \end{aligned}\end{equation*}
    then we could derive that
    \begin{equation}
    \begin{aligned}
    D(\beta_\ast)&=C(\beta_\ast)^{-1}\Cov_{P_\ast}\left(\frac{\partial L^{1}(\langle\mathbf{X},\beta\rangle,Y)}{\partial \beta}\right)\bigg\vert_{\beta=\beta_{\ast}} C(\beta_\ast)^{-1}\\
    &=\left(\mathbb{E}_{P_{\ast}}\left[ \frac{e^{\langle\mathbf{X},\beta_\ast\rangle}\mathbf{X}\mathbf{X}^{\top}}{\left( 1+e^{\langle\mathbf{X},\beta_{\ast}\rangle}\right)^2}\right]\right)^{-1}.
     \end{aligned}
    \end{equation}
   
    Regarding the asymptotic mean of $\beta_n^{ADRO}$, we have that
    \begin{equation}\label{h1}H(\beta_{\ast})=\tau\left(\sqrt{\mathbb{E}_{P_{\ast}}\left[ \frac{1}{\left(1+e^{Y\langle \mathbf{X},\beta_{\ast}\rangle}\right)^2}\right]}\frac{\beta_{\ast}}{\Vert\beta_{\ast}\Vert_2}-\frac{\Vert\beta_{\ast}\Vert_2 \mathbb{E}_{P_{\ast}}\left[ \frac{Ye^{Y\langle \mathbf{X},\beta_{\ast}\rangle}\mathbf{X}}{\left(1+e^{Y\langle \mathbf{X},\beta_{\ast}\rangle}\right)^3}\right]}{\sqrt{\mathbb{E}_{P_{\ast}}\left[ \frac{1}{\left(1+e^{Y\langle \mathbf{X},\beta_{\ast}\rangle}\right)^2}\right]}}\right).\end{equation}
    
    Notice we have that
    \begin{equation*}
    \begin{aligned}
    &\mathbb{E}_{P_{\ast}}\left[\frac{Ye^{Y\langle \mathbf{X},\beta_{\ast}\rangle}\mathbf{X}}{\left(1+e^{Y\langle \mathbf{X},\beta_{\ast}\rangle}\right)^3}\right]\\
    =&\int P_\ast(Y=1|\mathbf{X}=\mathbf{x}) \frac{e^{\langle \mathbf{x},\beta_{\ast}\rangle}\mathbf{x}}{\left(1+e^{\langle \mathbf{x},\beta_{\ast}\rangle}\right)^3}dF_\ast(\mathbf{x}) -\int P_\ast(Y=-1|\mathbf{X}=\mathbf{x}) \frac{e^{-\langle\mathbf{x},\beta_{\ast}\rangle}\mathbf{x}}{\left(1+e^{\langle \mathbf{x},\beta_{\ast}\rangle}\right)^3}dF_\ast(\mathbf{x})\\
    =&\int \frac{1}{1+e^{-\langle \mathbf{x},\beta_{\ast}\rangle}} \frac{e^{\langle \mathbf{x},\beta_{\ast}\rangle}\mathbf{x}}{\left(1+e^{\langle \mathbf{x},\beta_{\ast}\rangle}\right)^3}dF_\ast(\mathbf{x}) -\int \frac{1}{1+e^{\langle \mathbf{x},\beta_{\ast}\rangle}} \frac{e^{-\langle\mathbf{x},\beta_{\ast}\rangle}\mathbf{x}}{\left(1+e^{-\langle \mathbf{x},\beta_{\ast}\rangle}\right)^3}dF_\ast(\mathbf{x})\\
    =&\int \frac{1}{1+e^{\langle \mathbf{x},\beta_{\ast}\rangle}} \frac{e^{2\langle \mathbf{x},\beta_{\ast}\rangle}\mathbf{x}}{\left(1+e^{\langle \mathbf{x},\beta_{\ast}\rangle}\right)^3}dF_\ast(\mathbf{x}) -\int \frac{1}{1+e^{\langle \mathbf{x},\beta_{\ast}\rangle}} \frac{e^{2\langle\mathbf{x},\beta_{\ast}\rangle}\mathbf{x}}{\left(1+e^{\langle \mathbf{x},\beta_{\ast}\rangle}\right)^3}dF_\ast(\mathbf{x})\\
    =&\mathbf{0},
    \end{aligned}
    \end{equation*}
which indicates that the equation \eqref{sufficient2eq} holds and the second term in \eqref{h1} equals to $0$.

Then, we obtain that 
    \[H(\beta_{\ast})=\tau\sqrt{\mathbb{E}_{P_{\ast}}\left[ \frac{1}{\left(1+e^{Y\langle \mathbf{X},\beta_{\ast}\rangle}\right)^2}\right]}\frac{\beta_{\ast}}{\Vert\beta_{\ast}\Vert_2}.\]

    Notice we have that
    \begin{equation*}
    \begin{aligned}
    &\mathbb{E}_{P_{\ast}}\left[\frac{1}{\left(1+e^{Y\langle \mathbf{X},\beta_{\ast}\rangle}\right)^2}\right]\\
    =&\int P_\ast(Y=1|\mathbf{X}=\mathbf{x}) \frac{1}{\left(1+e^{\langle \mathbf{x},\beta_{\ast}\rangle}\right)^2}dF_\ast(\mathbf{x}) +\int P_\ast(Y=-1|\mathbf{X}=\mathbf{x}) \frac{1}{\left(1+e^{-\langle \mathbf{x},\beta_{\ast}\rangle}\right)^2}dF_\ast(\mathbf{x})\\
    =&\int \frac{1}{1+e^{-\langle \mathbf{x},\beta_{\ast}\rangle}} \frac{1}{\left(1+e^{\langle \mathbf{x},\beta_{\ast}\rangle}\right)^2}dF_\ast(\mathbf{x}) +\int \frac{1}{1+e^{\langle \mathbf{x},\beta_{\ast}\rangle}} \frac{1}{\left(1+e^{-\langle \mathbf{x},\beta_{\ast}\rangle}\right)^2}dF_\ast(\mathbf{x})\\
    =&\int \frac{1}{1+e^{\langle \mathbf{x},\beta_{\ast}\rangle}} \frac{e^{\langle \mathbf{x},\beta_{\ast}\rangle}}{\left(1+e^{\langle \mathbf{x},\beta_{\ast}\rangle}\right)^2}dF_\ast(\mathbf{x}) +\int \frac{1}{1+e^{\langle \mathbf{x},\beta_{\ast}\rangle}} \frac{e^{2\langle \mathbf{x},\beta_{\ast}\rangle}}{\left(1+e^{\langle \mathbf{x},\beta_{\ast}\rangle}\right)^2}dF_\ast(\mathbf{x})\\
    =&\int \frac{1}{1+e^{\langle \mathbf{x},\beta_{\ast}\rangle}} \frac{e^{\langle \mathbf{x},\beta_{\ast}\rangle}+e^{2\langle \mathbf{x},\beta_{\ast}\rangle}}{\left(1+e^{\langle \mathbf{x},\beta_{\ast}\rangle}\right)^2}dF_\ast(\mathbf{x})\\
    =&\mathbb{E}_{P_{\ast}}\left[\frac{e^{\langle\mathbf{X},\beta_{\ast}\rangle}}{\left(1+e^{\langle\mathbf{X},\beta_{\ast}\rangle}\right)^2}\right].
    \end{aligned}
    \end{equation*}

Then, $H(\beta_\ast)$ can be simplified as 
\begin{equation*}H(\beta_\ast)=\tau\sqrt{\mathbb{E}_{P_{\ast}}\left[\frac{e^{\langle\mathbf{X},\beta_{\ast}\rangle}}{\left(1+e^{\langle\mathbf{X},\beta_{\ast}\rangle}\right)^2}\right]}\frac{\beta_\ast}{\Vert\beta_\ast\Vert_2}.\end{equation*}
\end{proof}

\subsection{Proof of Proposition \ref{poissontheorem}}\label{proofpoisson}
\begin{proof}
Regarding the asymptotic covariance matrix,
    since we have that
    \begin{equation}\begin{aligned}
        &\Cov_{P_\ast}\left(\frac{\partial L^{2}(\langle\mathbf{X},\beta\rangle,Y)}{\partial \beta}\right)\bigg\vert_{\beta=\beta_{\ast}}\\
    =&\mathbb{E}_{P_\ast}\left[\frac{\partial L^{2}(\langle\mathbf{X},\beta\rangle,Y)}{\partial \beta}\left(\frac{\partial L^{2}(\langle\mathbf{X},\beta\rangle,Y)}{\partial \beta}\right)^\top\right]\Bigg\vert_{\beta=\beta_{\ast}}\\
    =&\mathbb{E}_{P_\ast}\left[ (e^{\langle\mathbf{X},\beta_\ast\rangle}-Y)^2\mathbf{X}\mathbf{X}^\top\right], \\=&\mathbb{E}_{P_{\ast}}\left[\mathbb{E}_{P_\ast}\left[ \left( e^{\langle \mathbf{X},\beta_\ast\rangle}-Y\right)^2\Big\vert \mathbf{X}\right]\mathbf{X}\mathbf{X}^\top\right]\\
    =&\mathbb{E}_{P_{\ast}}\left[ \Var_{P_\ast}(Y\vert\mathbf{X})\mathbf{X}\mathbf{X}^{\top}\right]\\
=&\mathbb{E}_{P_{\ast}}\left[ e^{\langle \mathbf{X},\beta_{\ast}\rangle}\mathbf{X}\mathbf{X}^{\top}\right],\end{aligned}\end{equation}
and
\[C(\beta_\ast)=\mathbb{E}_{P_\ast}\left[ \frac{\partial^2 L^{2}(\langle\mathbf{X},\beta\rangle,Y)}{\partial \beta^2}\right]\bigg\vert_{\beta=\beta_\ast}=\mathbb{E}_{P_{\ast}}\left[ e^{\langle \mathbf{X},\beta_{\ast}\rangle}\mathbf{X}\mathbf{X}^{\top}\right],\]
   then we could derive that
    \begin{equation}\begin{aligned}
    D(\beta_\ast)&=C(\beta_\ast)^{-1}\Cov_{P_\ast}\left(\frac{\partial L^{2}(\langle\mathbf{X},\beta\rangle,Y)}{\partial \beta}\right)\bigg\vert_{\beta=\beta_{\ast}}C(\beta_\ast)^{-1}\\
    &=\left(\mathbb{E}_{P_{\ast}}\left[ e^{\langle \mathbf{X},\beta_{\ast}\rangle}\mathbf{X}\mathbf{X}^{\top}\right]\right)^{-1}.    \end{aligned}\end{equation}
   
    Regarding the asymptotic mean of $\beta_n^{ADRO}$, we have that
\begin{equation}\label{h2}H(\beta_\ast)=\tau \left(\sqrt{\mathbb{E}_{P_{\ast}}\left[\left( e^{\langle\mathbf{X},\beta_\ast\rangle}-Y\right)^2\right]} \frac{\beta_\ast}{\Vert\beta_\ast\Vert_2}-\Vert\beta_\ast\Vert_2\frac{\mathbb{E}_{P_{\ast}}\left[ \left(e^{\langle \mathbf{X},\beta_\ast\rangle}-Y\right)e^{\langle\mathbf{X},\beta_\ast\rangle}\mathbf{X}\right]}{\sqrt{\mathbb{E}_{P_{\ast}}\left[\left( e^{\langle\mathbf{X},\beta_\ast\rangle}-Y\right)^2\right]}}\right),\end{equation}
For the second term, we have that
\begin{equation*}
\begin{aligned}
&\mathbb{E}_{P_{\ast}}\left[ \left(e^{\langle \mathbf{X},\beta_\ast\rangle}-Y\right)e^{\langle\mathbf{X},\beta_\ast\rangle}\mathbf{X}\right]\\
=&\mathbb{E}_{P_{\ast}}\left[e^{\langle\mathbf{X},\beta_\ast\rangle}\mathbb{E}_{P_{\ast}}\left[e^{\langle \mathbf{X},\beta_\ast\rangle}-Y\big\vert\mathbf{X}\right]\mathbf{X}\right]\\
=& \mathbf{0},
\end{aligned}
\end{equation*}
which indicates that the equation \eqref{sufficient2eq} holds and the second term in \eqref{h2} equals to $0$.

Further, we have that
\begin{equation*}
\begin{aligned}
&\mathbb{E}_{P_{\ast}}\left[\left( e^{\langle\mathbf{X},\beta_\ast\rangle}-Y\right)^2\right]\\
=&\mathbb{E}_{P_{\ast}}\left[\mathbb{E}_{P_{\ast}}\left[\left( e^{\langle\mathbf{X},\beta_\ast\rangle}-Y\right)^2\Big\vert\mathbf{X}\right]\right]\\
=&\mathbb{E}_{P_{\ast}}\left[\Var_{P_\ast}(Y\vert\mathbf{X})\right]\\
=&\mathbb{E}_{P_{\ast}}\left[e^{\langle\mathbf{X},\beta_\ast\rangle}\right].
\end{aligned}
\end{equation*}

 Hence, we have that
\[H(\beta_{\ast})=\tau\sqrt{\mathbb{E}_{P_\ast}[e^{\langle \mathbf{X},\beta_{\ast}\rangle}]}\frac{\beta_\ast}{\Vert\beta_\ast\Vert_2}.\]
\end{proof}
\subsection{Proof of Proposition \ref{lineartheorem}}\label{prooflinear}
\begin{proof}
Regarding the asymptotic covariance matrix,
    since we have that
    \begin{equation*}
    \begin{aligned}
    \Cov_{P_\ast}\left(\frac{\partial L^{3}(\langle\mathbf{X},\beta\rangle,Y)}{\partial \beta}\right)\bigg\vert_{\beta=\beta_{\ast}}&=\mathbb{E}_{P_\ast}\left[\frac{\partial L^{3}(\langle\mathbf{X},\beta\rangle,Y)}{\partial \beta}\left(\frac{\partial L^{3}(\langle\mathbf{X},\beta\rangle,Y)}{\partial \beta}\right)^\top\right]\Bigg\vert_{\beta=\beta_{\ast}}\\
    &=\mathbb{E}_{P_\ast}\left[  (\langle\mathbf{X},\beta_\ast\rangle-Y)^2\mathbf{X}\mathbf{X}^\top\right]\\
    &=\mathbb{E}_{P_\ast}\left[\mathbb{E}_{P_\ast}\left[ (\langle\mathbf{X},\beta_\ast\rangle-Y)^2\vert \mathbf{X}\right]\mathbf{X}\mathbf{X}^\top\right]\\
    &= \mathbb{E}_{P_\ast}\left[\Var_{P_\ast}(Y\vert\mathbf{X})\mathbf{X}\mathbf{X}^\top\right]\\
    &=\sigma^2 \mathbb{E}_{P_{\ast}}\left[ \mathbf{X}\mathbf{X}^{\top}\right],
     \end{aligned}\end{equation*}
     and
      \[C=\mathbb{E}_{P_\ast}\left[\frac{\partial^2 L^{3}(\langle\mathbf{X},\beta\rangle,Y)}{\partial \beta^2}\right]=\mathbb{E}_{P_{\ast}}\left[ \mathbf{X}\mathbf{X}^{\top}\right],\]
    then we could derive that
    \begin{equation*}\begin{aligned}     
  D&=C^{-1}\Cov_{P_\ast}\left(\frac{\partial L^{1}(\langle\mathbf{X},\beta\rangle,Y)}{\partial \beta}\right)\Bigg\vert_{\beta=\beta_{\ast}} C^{-1}\\
  &=\sigma^2\left(\mathbb{E}_{P_{\ast}}\left[ \mathbf{X}\mathbf{X}^{\top}\right]\right)^{-1}.  \end{aligned}\end{equation*}
  
    Regarding the asymptotic mean of $\beta_n^{ADRO}$, it follows from \eqref{decomposeH} that 
    \begin{equation}\label{h3}H(\beta_{\ast})=\tau\left(\sqrt{\mathbb{E}_{P_{\ast}}\left[ (\langle\mathbf{X},\beta_\ast\rangle-Y)^2\right]}\frac{\beta_{\ast}}{\Vert\beta_{\ast}\Vert_2}-\frac{\Vert\beta_{\ast}\Vert_2 \mathbb{E}_{P_{\ast}}\left[ (\langle\mathbf{X},\beta_\ast\rangle-Y)\mathbf{X}\right]}{\sqrt{\mathbb{E}_{P_{\ast}}\left[ (\langle\mathbf{X},\beta_\ast\rangle-Y)^2\right]}}\right).\end{equation}
    
    For the second term, we have that
\begin{equation*} \begin{aligned}
    &\mathbb{E}_{P_{\ast}}\left[ (\langle\mathbf{X},\beta_\ast\rangle-Y)\mathbf{X}\right]\\
    =& \mathbb{E}_{P_\ast}\left[\langle\mathbf{X},\beta_\ast\rangle\mathbf{X}-\mathbb{E}_{P_{\ast}}\left[Y\vert\mathbf{X}\right]\mathbf{X}\right]\\
    =& \mathbf{0},
    \end{aligned}\end{equation*}
indicating that the equation \eqref{sufficient2eq} holds and the second term in \eqref{h3} equals to $0$.

Then, we obtain that 
    \[H(\beta_{\ast})=\tau\sqrt{\mathbb{E}_{P_{\ast}}\left[ (\langle\mathbf{X},\beta_\ast\rangle-Y)^2\right]}\frac{\beta_{\ast}}{\Vert\beta_{\ast}\Vert_2}.\]
  
Notice  we also have that
\begin{equation*}
\begin{aligned}&\mathbb{E}_{P_{\ast}}\left[ (\langle\mathbf{X},\beta_\ast\rangle-Y)^2\right]\\
=&\mathbb{E}_{P_{\ast}}\left[\mathbb{E}_{P_\ast}\left[\left( \langle \mathbf{X},\beta_\ast\rangle-Y\right)^2\vert \mathbf{X}\right]\right]\\
=&\mathbb{E}_{P_\ast}\left[\Var_{P_\ast}(Y\vert\mathbf{X})\right]\\
=&\sigma^2.
\end{aligned}
\end{equation*} 

Thus, we have that 
\[H(\beta_{\ast})=\tau\sigma\frac{\beta_{\ast}}{\Vert\beta_\ast\Vert_2}.\]
\end{proof}
\newpage
\bibliography{sample}
\end{document}